\documentclass[10pt,journal,compsoc]{IEEEtran}

\ifCLASSOPTIONcompsoc
  \usepackage{cite}
\else
  \usepackage{cite}
\fi

\ifCLASSINFOpdf
  \usepackage[pdftex]{graphicx}
\else
  \usepackage[dvips]{graphicx}
\fi

\usepackage{amsmath}
\usepackage{amssymb}
\usepackage{amsfonts} 

\ifCLASSOPTIONcompsoc
  \usepackage[caption=false,font=footnotesize,labelfont=sf,textfont=sf]{subfig}
\else
  \usepackage[caption=false,font=footnotesize]{subfig}
\fi

\usepackage{url}
\usepackage{color}
\usepackage{multirow}
\usepackage{hyperref}
\usepackage{amsmath}
\usepackage{amssymb}
\usepackage{colortbl}

\usepackage{etoolbox}
\makeatletter
\patchcmd{\@makecaption}
  {\scshape}
  {}
  {}
  {}
\makeatother

\usepackage{pifont}% http://ctan.org/pkg/pifont

\newcommand{\etal}{\textit{et al}.}
\newcommand{\ie}{\textit{i}.\textit{e}.}
\newcommand{\eg}{\textit{e}.\textit{g}.}

\begin{document}
%
% paper title
% Titles are generally capitalized except for words such as a, an, and, as,
% at, but, by, for, in, nor, of, on, or, the, to and up, which are usually
% not capitalized unless they are the first or last word of the title.
% Linebreaks \\ can be used within to get better formatting as desired.
% Do not put math or special symbols in the title.
\title{Self-Supervised Masked Convolutional Transformer Block for Anomaly Detection}

\author{Neelu~Madan,
        {Nicolae-C\u{a}t\u{a}lin}~Ristea,
        {Radu Tudor}~Ionescu,~\IEEEmembership{Member,~IEEE,}
        Kamal~Nasrollahi,\\
        {Fahad Shahbaz}~Khan,~\IEEEmembership{Senior Member,~IEEE,}
        {Thomas B.}~Moeslund,
        and~Mubarak~Shah,~\IEEEmembership{Fellow,~IEEE}

       % and~Marius~Popescu% <-this % stops a space
\IEEEcompsocitemizethanks{\IEEEcompsocthanksitem N. Madan is with the Center for Research in Computer Vision (CRCV), Department of Computer Science, University of Central Florida, Orlando, FL, 32816 and the Department of Architecture, Design, and Media Technology, Aalborg University, Denmark.
\IEEEcompsocthanksitem N.C. Ristea is with the Department of Telecommunications, University Politehnica of Bucharest, Romania, and the Department of Computer Science, University of Bucharest, Romania.
\IEEEcompsocthanksitem R.T. Ionescu is with SecurifAI, the Romanian Young Academy, and the Department of Computer Science, University of Bucharest, Romania. R.T. Ionescu is the corresponding author.\protect\\
% note need leading \protect in front of \\ to get a newline within \thanks as
% \\ is fragile and will error, could use \hfil\break instead.
E-mail: raducu.ionescu@gmail.com
\IEEEcompsocthanksitem K. Nasrollahi is with the Department of Architecture, Design, and Media Technology, Aalborg University, Denmark, and Milestone Systems, Denmark.
\IEEEcompsocthanksitem F.S. Khan is with Mohamed bin Zayed University of Artificial Intelligence (MBZUAI), UAE, and
Link\"{o}ping University, Sweden.
\IEEEcompsocthanksitem Thomas B. Moeslund is with the Department of Architecture, Design, and Media Technology, Aalborg University, Denmark.
\IEEEcompsocthanksitem M. Shah is with the Center for Research in Computer Vision (CRCV), Department of Computer Science, University of Central Florida, Orlando, FL, 32816.
\IEEEcompsocthanksitem N. Madan and N.C. Ristea contributed equally.}
% <-this % stops an unwanted space
\thanks{Manuscript received April 19, 2022; revised August 26, 2022.}}

% note the % following the last \IEEEmembership and also \thanks - 
% these prevent an unwanted space from occurring between the last author name
% and the end of the author line. i.e., if you had this:
% 
% \author{....lastname \thanks{...} \thanks{...} }
%                     ^------------^------------^----Do not want these spaces!
%
% a space would be appended to the last name and could cause every name on that
% line to be shifted left slightly. This is one of those "LaTeX things". For
% instance, "\textbf{A} \textbf{B}" will typeset as "A B" not "AB". To get
% "AB" then you have to do: "\textbf{A}\textbf{B}"
% \thanks is no different in this regard, so shield the last } of each \thanks
% that ends a line with a % and do not let a space in before the next \thanks.
% Spaces after \IEEEmembership other than the last one are OK (and needed) as
% you are supposed to have spaces between the names. For what it is worth,
% this is a minor point as most people would not even notice if the said evil
% space somehow managed to creep in.

% The paper headers
\markboth{IEEE Transactions on Pattern Analysis and Machine Intelligence,~Vol.~14, No.~8, August~2022}%
{Madan \MakeLowercase{\textit{et al.}}: Self-Supervised Masked Convolutional Transformer Block for Anomaly Detection}

% The publisher's ID mark at the bottom of the page is less important with
% Computer Society journal papers as those publications place the marks
% outside of the main text columns and, therefore, unlike regular IEEE
% journals, the available text space is not reduced by their presence.
% If you want to put a publisher's ID mark on the page you can do it like
% this:
%\IEEEpubid{0000--0000/00\$00.00~\copyright~2015 IEEE}
% or like this to get the Computer Society new two part style.
%\IEEEpubid{\makebox[\columnwidth]{\hfill 0000--0000/00/\$00.00~\copyright~2015 IEEE}%
%\hspace{\columnsep}\makebox[\columnwidth]{Published by the IEEE Computer Society\hfill}}
% Remember, if you use this you must call \IEEEpubidadjcol in the second
% column for its text to clear the IEEEpubid mark (Computer Society jorunal
% papers don't need this extra clearance.)

\IEEEtitleabstractindextext{%
\begin{abstract}
Anomaly detection has recently gained increasing attention in the field of computer vision, likely due to its broad set of applications ranging from product fault detection on industrial production lines and impending event detection in video surveillance to finding lesions in medical scans. Regardless of the domain, anomaly detection is typically framed as a one-class classification task, where the learning is conducted on normal examples only. An entire family of successful anomaly detection methods is based on learning to reconstruct masked normal inputs (\eg~patches, future frames, etc.) and exerting the magnitude of the reconstruction error as an indicator for the abnormality level. Unlike other reconstruction-based methods, we present a novel self-supervised masked convolutional transformer block (SSMCTB) that comprises the reconstruction-based functionality at a core architectural level. The proposed self-supervised block is extremely flexible, enabling information masking at any layer of a neural network and being compatible with a wide range of neural architectures. In this work, we extend our previous self-supervised predictive convolutional attentive block (SSPCAB) with a 3D masked convolutional layer, a transformer for channel-wise attention, as well as a novel self-supervised objective based on Huber loss. Furthermore, we show that our block is applicable to a wider variety of tasks, adding anomaly detection in medical images and thermal videos to the previously considered tasks based on RGB images and surveillance videos. We exhibit the generality and flexibility of SSMCTB by integrating it into multiple state-of-the-art neural models for anomaly detection, bringing forth empirical results that confirm considerable performance improvements on five benchmarks: MVTec AD, BRATS, Avenue, ShanghaiTech, and Thermal Rare Event. We release our code and data as open source at: \url{https://github.com/ristea/ssmctb}.
\end{abstract}
% Note that keywords are not normally used for peerreview papers.
}

\maketitle

\IEEEdisplaynontitleabstractindextext

% For peerreview papers, this IEEEtran command inserts a page break and
% creates the second title. It will be ignored for other modes.
\IEEEpeerreviewmaketitle

\IEEEraisesectionheading{\section{Introduction}\label{sec:introduction}}

\IEEEPARstart{T}{he} applications of vision-based anomaly detection are very diverse, ranging from industrial settings, where the need is to detect faulty objects in the production line \cite{Bergmann-CVPR-2019,Lee-A-2022}, to video surveillance, where the need is to detect abnormal behavior \cite{Luo-ICCV-2017} such as people fighting or shoplifting, and even medical imaging, where the need is to detect abnormal tissue \cite{Shvetsova-A-2021} such as malignant lesions. One of the major challenges of the anomaly detection task is that the definition of what represents an anomaly implies a high dependence on context. For instance, a car driven in a pedestrian area is labeled as anomalous, whereas the same action can be considered normal in a different context, \eg~when the car is driven on the road. Due to the reliance on context and the sheer diversity of possible anomalies, it is often very difficult to gather abnormal examples for training. As a result, anomaly detection is commonly devised as a one-class classification task, where the generic approach implicitly or explicitly learns the distribution of the normal training data. During inference, examples that do not belong to the normal training data distribution are labeled as abnormal. There are several categories of methods that are guided by this generic approach, such as dictionary-learning methods \cite{Carrera-TII-2017,Cheng-CVPR-2015,Cong-CVPR-2011, Dutta-AAAI-2015,Lu-ICCV-2013,Ren-BMVC-2015}, change-detection frameworks \cite{Giorno-ECCV-2016,Ionescu-ICCV-2017,Liu-BMVC-2018,Pang-CVPR-2020}, distance-based models \cite{Bergmann-CVPR-2020,Defard-ICPR-2021,Ionescu-CVPR-2019,Ionescu-WACV-2019,Ramachandra-WACV-2020a,Ramachandra-WACV-2020b,Ravanbakhsh-WACV-2018,Sabokrou-IP-2017,Sabokrou-CVIU-2018,Saligrama-CVPR-2012,Smeureanu-ICIAP-2017,Sun-PR-2017,Tran-BMVC-2017}, probabilistic frameworks \cite{Adam-PAMI-2008,Antic-ICCV-2011,Feng-NC-2017,Hinami-ICCV-2017,Kim-CVPR-2009,Mahadevan-CVPR-2010,Mehran-CVPR-2009,Rudolph-WACV-2021,Saleh-CVPR-2013,Wu-CVPR-2010}, and reconstruction-based models \cite{Fei-TMM-2020,Gong-ICCV-2019,Hasan-CVPR-2016,Li-BMVC-2020,Liu-CVPR-2018,Luo-ICCV-2017,Nguyen-ICCV-2019,Park-CVPR-2020,Ravanbakhsh-ICIP-2017,Salehi-CVPR-2021,Tang-PRL-2020,Venkataramanan-ECCV-2020}. 

\begin{figure*}[!t]
\begin{center}
\centerline{\includegraphics[width=1.0\linewidth]{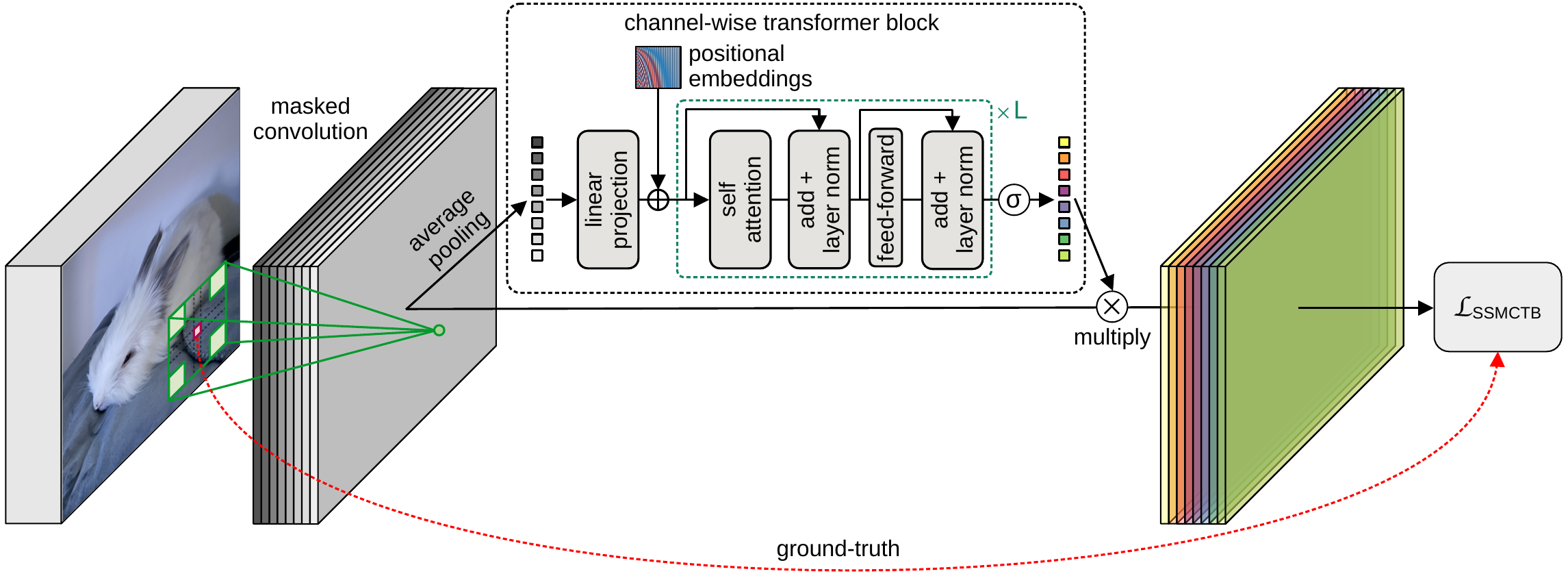}}
\vspace{-0.25cm}
\caption{An overview of our self-supervised masked convolutional transformer block (SSMCTB). At every location where the masked filters are applied, the proposed block has to rely on the visible regions (sub-kernels) to reconstruct the masked region (center area). A transformer module performs channel-wise self-attention to selectively promote or suppress reconstruction maps via a set of weights returned by a sigmoid ($\sigma$) layer. The block is self-supervised via the Huber loss ($\mathcal{L}_{\mbox{\tiny{SSMCTB}}}$) \cite{Huber-AMS-1964} between masked and returned activation maps. Best viewed in color.}
\label{fig_sspcab}
%\vspace{-0.9cm}
\end{center}
\end{figure*}

Our approach belongs to the category of reconstruction methods, which have recently become a prominent choice in anomaly detection \cite{Fei-TMM-2020,Gong-ICCV-2019,Li-BMVC-2020,Nguyen-ICCV-2019,Park-CVPR-2020,Salehi-CVPR-2021,Tang-PRL-2020,Venkataramanan-ECCV-2020}. Reconstruction-based models implicitly learn the normal data distribution by minimizing the reconstruction error of the normal instances at training time. These models are based on the assumption that the learned latent manifold does not offer the means to reconstruct the abnormal samples robustly, due to the unavailability of such samples at training time. Hence, the reconstruction error is directly employed as the anomaly score. 
A particular subcategory of reconstruction-based models relies on learning to predict masked inputs \cite{Georgescu-CVPR-2021,Acsintoae-CVPR-2022,Li-BMVC-2020,Liu-CVPR-2018} as a self-supervised pretext task. In this case, the reconstruction error with respect to the masked information is used to assess the abnormality level of an input instance. Depending on the input type (image or video), methods in this subcategory mask various parts of the input, \eg~superpixels in images \cite{Li-BMVC-2020}, future frames in video \cite{Liu-CVPR-2018}, or middle bounding boxes in object-centric temporal sequences \cite{Georgescu-CVPR-2021,Acsintoae-CVPR-2022}, and employ the whole model to reconstruct the masked input. We, on the other hand, propose to encapsulate the functionality of reconstructing the masked information into a novel neural block. There are two major benefits when wrapping the reconstruction task as a low-level architectural component: $(i)$ it enables introducing the reconstruction of masked information as a self-supervised task at any layer of a neural network (not only at the input), and $(ii)$ it eases integrating the self-supervised reconstruction task into a broad variety of neural architectures, regardless of whether the respective models are reconstruction-based or not. Due to its advantages, our block is very flexible and generic.

Our self-supervised reconstruction block consists of a dilated masked convolution followed by a channel-wise transformer module. The center area of our convolutional kernel is masked, hence hiding the center of the receptive field at every location where the filters are applied. In other words, each component of the input tensor is certainly masked at some point during the convolution operation, which means that the entire input tensor ends up being masked. Next, the convolutional activation maps are transformed into tokens using an average pooling layer. Then, the resulting tokens are passed through a transformer module \cite{Dosovitskiy-ICLR-2020,Vaswani-NIPS-2017} that performs channel-wise self-attention. The proposed block is equipped with a transformer module to avoid the direct reconstruction of the masked area through linearly interpolating the visible regions of the convolutional kernels. The final activation maps are multiplied with the resulting attention tokens. Our block is designed in such a way that the output tensor has the same dimensions as the input tensor, which allows us to easily introduce a loss within our block to minimize the reconstruction error between the output tensor and the masked input tensor. By integrating this loss, our block becomes a self-contained trainable component that learns to predict the masked information via self-supervision. As such, we coin the term \emph{self-supervised masked convolutional transformer block} (SSMCTB) to designate our novel neural component for anomaly detection. As shown in Figure~\ref{fig_sspcab}, SSMCTB learns to reconstruct the masked region based on the available context (visible regions of the receptive field), for each location where the dilated kernels are applied. Notably, we can graciously control the level (from local to global) of the contextual information by choosing the appropriate dilation rate for the masked kernels. 

SSMCTB is an extension of the self-supervised predictive convolutional attentive block (SSPCAB) introduced in our recent CVPR 2022 paper \cite{Ristea-CVPR-2022}. In the current work, we modify SSPCAB in three different ways: $(i)$ we replace the standard channel attention module in the original SSPCAB \cite{Ristea-CVPR-2022} with a multi-head self-attention module \cite{Dosovitskiy-ICLR-2020,Vaswani-NIPS-2017} to increase the modeling capacity, $(ii)$ we extend the masked convolution operation with 3D convolutional filters, enabling the integration of SSMCTB into networks based on 3D convolutional layers, and $(iii)$ we replace the mean squared error (MSE) loss with the Huber loss \cite{Huber-AMS-1964}, since the latter loss is less sensitive to outliers than the former loss. Aside from these architectural changes, we demonstrate the applicability of our block to more domains, adding anomaly detection in medical images and thermal videos to the previously considered tasks based on RGB images and surveillance videos. Moreover, we conduct a more extensive ablation study, thus providing a more comprehensive set of results. We also show that our module is suitable for both convolutional and transformer-based architectures.

We introduce SSMCTB into multiple state-of-the-art neural models \cite{Barbalau-ARXIV-2022, Zavrtanik-ICCV-2021, Schulter-ECCV-2022, Georgescu-TPAMI-2021, Liu-ICCV-2021, Park-CVPR-2020, Liu-CVPR-2018, He-CVPR-2022,Vasu-ICCV-2023,Wang-ICDM-2022} for anomaly detection and conduct experiments on five benchmarks: MVTec AD \cite{Bergmann-CVPR-2019}, BRATS \cite{Menze-TMI-2015}, Avenue \cite{Lu-ICCV-2013}, ShanghaiTech \cite{Luo-ICCV-2017}, and Thermal Rare Event. The Thermal Rare Event data set is a novel benchmark for anomaly detection, which we constructed by manually labeling abnormal events from the Seasons in Drift data set \cite{Nikolov-NIPS-2021}. The chosen benchmarks belong to various domains, ranging from industrial and medical images to RGB and thermal videos. This is to show that SSMCTB is applicable to multiple domains. When adding SSMCTB to the state-of-the-art models, our experiments show evidence of consistent improvements across all models and tasks, indicating that our block is generic and easily adaptable. When compared to SSPCAB, we observe performance gains in the majority of cases, showing that the multi-head self-attention and the Huber loss are beneficial in detriment of the standard channel attention \cite{Hu-CVPR-2018} and the MSE loss, respectively.

In summary, our contribution is multifold:
\begin{itemize}
    \item We introduce the masked convolution operation and integrate it into a novel self-supervised masked convolutional transformer block which exhibits the inherent ability to detect anomalies.
    \item We encapsulate our block into several state-of-the-art methods \cite{Barbalau-ARXIV-2022, Zavrtanik-ICCV-2021, Schulter-ECCV-2022, Georgescu-TPAMI-2021, Liu-ICCV-2021, Park-CVPR-2020, Liu-CVPR-2018, He-CVPR-2022,Vasu-ICCV-2023,Wang-ICDM-2022} for anomaly detection, showing considerable performance gains across multiple models, benchmarks and domains.
    \item We extend the 2D masked convolution to a 3D masked convolution that considers a 3D context, and we integrate the new 3D SSMCTB into two 3D networks for anomaly detection \cite{Barbalau-ARXIV-2022, Zavrtanik-ICCV-2021}.
    \item We replace the Squeeze-and-Excitation module \cite{Hu-CVPR-2018} of SSPCAB with a transformer module that performs channel-wise attention.
    \item We substitute the MSE loss in SSPCAB with the Huber loss, improving the sensitivity to outliers during self-supervised learning.
    \item We conduct a more comprehensive set of experiments, including new methods and benchmarks from previously missing domains (medical images, thermal videos).
    \item We provide an extensive ablation study, including different variations of the proposed self-supervised block.
    \item We annotate a subset (one week of video) of the Seasons in Drift \cite{Nikolov-NIPS-2021} data set with anomaly labels, obtaining a new benchmark for anomaly detection in thermal videos.
\end{itemize}

% Proposed a thermal dataset for anomaly detection
% The masked convolution is providing consistent improvement across the cross domain. 

\section{Related Work}

\subsection{Transformers}

Vaswani \etal~\cite{Vaswani-NIPS-2017} introduced the self-attention mechanism, sparking the research of neural architectures relying solely on attention, including research on vision transformers \cite{Carion-ECCV-2020,Chen-arXiv-2021,Dosovitskiy-ICLR-2020,Khan-ACS-2021,Parmar-ICML-2018,Ristea-A-2021,Touvron-ICML-2021,Wu-ICCV-2021,Xu-AAAI-2022,Zhang-CVPR-2022,Zheng-BMVC-2021,Zhu-ICLR-2020}. These models are now embraced at a fast pace in the field of computer vision, certainly due to the imposing performance levels across a broad variety of tasks, ranging from object recognition \cite{Dosovitskiy-ICLR-2020,Touvron-ICML-2021,Wu-ICCV-2021} and object detection \cite{Carion-ECCV-2020,Zheng-BMVC-2021,Zhu-ICLR-2020} to image generation \cite{Ristea-A-2021,Xu-AAAI-2022,Zhang-CVPR-2022} and anomaly detection \cite{Wang-ICDM-2022,Jiang-TII-2023,Lee-A-2022b,Mishra-ISIE-2021,Pirnay-ICIAP-2022}. Unlike approaches using only transformer-based attention \cite{Carion-ECCV-2020,Chen-arXiv-2021,Dosovitskiy-ICLR-2020,Khan-ACS-2021,Parmar-ICML-2018,Ristea-A-2021,Touvron-ICML-2021,Wu-ICCV-2021,Zheng-BMVC-2021,Zhu-ICLR-2020,Xu-ICLR-2022}, we propose a novel and flexible block that employs transformer-based attention along with masked convolution, which can be integrated into multiple architectures that are not necessarily transformer-based. To endorse this statement, we introduce SSMCTB into a variety of models and conduct a series of experiments showing that our block can bring significant performance gains. Another difference from vision transformers is that our block performs channel-wise self-attention, while conventional vision transformers perform spatial attention \cite{Dosovitskiy-ICLR-2020}. We conduct an ablation study to compare channel and spatial attention inside SSMCTB, showing that channel attention provides superior performance and faster processing.

\subsection{Self-Supervision via Information Masking} 

The reconstruction of masked information has recently become an attractive area of interest \cite{He-CVPR-2022, Pathak-CVPR-2016, Wei-CVPR-2022, Chang-CVPR-2022, Yu-CVPR-2022}. Models based on information masking are usually pre-trained on a self-supervised reconstruction task, being later employed for downstream visual tasks such as object detection and image segmentation. For instance, He \etal~\cite{He-CVPR-2022} proposed to reconstruct masked (erased) patches as a self-supervised pretext task for pre-training auto-encoders, subsequently using them for mainstream tasks, including object detection and object recognition. They reported optimal results when a majority (75\%) of the patches is masked. Masked auto-encoders are directly applicable to anomaly detection. However, we show that SSMCTB can boost the performance of masked auto-encoders, suggesting that it can leverage information masking in a distinct way. Wei \etal~\cite{Wei-CVPR-2022} aimed at pre-training video models, proposing to mask spatio-temporal cubes from a video and predict the features of the masked regions. Chang \etal~\cite{Chang-CVPR-2022} introduced a bidirectional decoder that learns to predict masked tokens by attending them from all directions. The proposed method provides an efficient substitute for generative transformers. 
Yu \etal~\cite{Yu-CVPR-2022} used a masked point modeling task for pre-training a point cloud transformer. They showed that the representation learned by the model transfers well to new (downstream) tasks and domains. Distinct from such methods, we integrate information masking at a core operational level inside neural networks via our masked convolutional layer. We self-supervise our block (which incorporates masked convolution) through a reconstruction loss and show that modeling the context towards reconstructing the masked information results in an effective discriminative manifold for anomaly detection.

We underline that some recent approaches \cite{Georgescu-CVPR-2021, Astrid-ICCVW-2021, Liu-CVPR-2018} utilize masking as a surrogate task for anomaly detection. We discuss these methods and explain how our approach is different in a separate subsection below. 

\subsection{Anomaly Detection}

Anomaly detection frameworks are usually trained in a one-class setting, where only normal data is available at training time, whereas both normal and abnormal examples are present at test time. The anomaly detection methods operating in this setting can be classified into different categories, which are briefly presented below. Dictionary learning methods \cite{Carrera-TII-2017,Cheng-CVPR-2015,Cong-CVPR-2011,Dutta-AAAI-2015,Lu-ICCV-2013,Ren-BMVC-2015} construct a dictionary of atoms from normal instances, labeling examples that are not represented in the dictionary as abnormal. Change detection frameworks \cite{Giorno-ECCV-2016,Ionescu-ICCV-2017,Liu-BMVC-2018,Pang-CVPR-2020} are applied directly on test videos, measuring the degree of change between current and preceding video frames to detect anomalies. Probabilistic models \cite{Adam-PAMI-2008,Antic-ICCV-2011,Feng-NC-2017,Hinami-ICCV-2017,Kim-CVPR-2009,Mahadevan-CVPR-2010,Mehran-CVPR-2009,Rudolph-WACV-2021,Saleh-CVPR-2013,Wu-CVPR-2010} learn the probability density function of the normal data, flagging examples outside the distribution as abnormal. Distance-based approaches \cite{Bergmann-CVPR-2020,Defard-ICPR-2021,Ionescu-CVPR-2019,Ionescu-WACV-2019,Ramachandra-WACV-2020a,Ramachandra-WACV-2020b,Ravanbakhsh-WACV-2018,Sabokrou-IP-2017,Sabokrou-CVIU-2018,Saligrama-CVPR-2012,Smeureanu-ICIAP-2017,Sun-PR-2017,Tran-BMVC-2017,Roth-CVPR-2022} learn a distance function between samples, such that the distance between normal instances is lower than the distance between normal and abnormal instances. Reconstruction-based methods \cite{Fei-TMM-2020,Gong-ICCV-2019,Hasan-CVPR-2016,Li-BMVC-2020,Liu-CVPR-2018,Luo-ICCV-2017,Nguyen-ICCV-2019,Park-CVPR-2020,Ravanbakhsh-ICIP-2017,Salehi-CVPR-2021,Tang-PRL-2020,Venkataramanan-ECCV-2020, Zavrtanik-ICCV-2021,Yamada-IROS-2022} learn to reconstruct normal examples, detecting anomalies based on the magnitude of the reconstruction error, as anomalies tend to have larger errors than normal instances. 

% Start Explaining reconstruction based methods here (13/07)
\noindent
\textbf{Reconstruction-based methods.} 
Since our block belongs to the category of reconstruction-based models, we discuss this category in more detail next.
Reconstruction-based models are often chosen for both image and video anomaly detection \cite{Zavrtanik-ICCV-2021, Georgescu-CVPR-2021, Park-CVPR-2020, Liu-ICCV-2021}. These approaches typically employ auto-encoders and generative adversarial networks (GANs) to learn a powerful latent manifold representing the normal data distribution. For the video domain, some anomaly detection approaches \cite{Liu-ICCV-2021, Ionescu-CVPR-2019, Liu-CVPR-2018} incorporate additional cues by reconstructing the optical flow to capture motion information, enabling the detection of motion-based anomalies such as running and jumping. Doshi \etal~\cite{Doshi-WACV-2022} proposed a continual learning setup, which could be easily extended for future normal and abnormal patterns.  

As the amount of normal training data is generally high, latent manifolds show a tendency to generalize too well, being capable of reconstructing  abnormal instances with low error. In the context of anomaly detection, generalizing to out-of-distribution samples, \eg~anomalies, is not desired, although this would be mostly desirable in other application domains. To mitigate this issue, researchers employed various techniques, such as adding memory modules \cite{Gong-ICCV-2019,Liu-ICCV-2021,Park-CVPR-2020} or pseudo-anomalies during training \cite{Astrid-ICCVW-2021,Georgescu-TPAMI-2021}. Memory-based auto-encoders \cite{Gong-ICCV-2019,Liu-ICCV-2021} generally employ an additional module to memorize the normal patterns observed in the training data. Consequently, memory modules increase the computational complexity of the model, and the faithful reconstruction of normal samples highly relies on the size of the memory module. Georgescu \etal~\cite{Georgescu-TPAMI-2021} proposed to optimize the model on pseudo-anomalies with gradient ascent, while still using gradient descent to learn the normal data distribution. This results in a powerful discriminative subspace for the robust detection of the abnormal samples. The pseudo-abnormal instances are samples collected from different contexts, such as flowers, animals, cartoons, and textures, unrelated to the object distribution (comprising humans, cars, bicycles, etc.) observed in typical urban surveillance scenes. Similarly, Astrid \etal~\cite{Astrid-ICCVW-2021} generated pseudo-anomalies by skipping a few frames from the video and training an auto-encoder by maximizing the loss for pseudo-anomalies and minimizing it for normal samples. Introducing pseudo-anomalies increases the training time and may sometimes cause instability if the balance between gradient descent on normal data and gradient ascent on pseudo-abnormal data is not tuned. Different from related reconstruction-based methods, we increase the difficulty of the reconstruction task by masking information wherever SSMCTB is introduced into a neural model, thus making it harder for the model to generalize to abnormal data. As shown by our experimental results, our block adds a marginal computational overhead.

% reconstruction of masked information in anomaly detection (can be expanded more)
\noindent
\textbf{Masking for Anomaly Detection.}
Some approaches \cite{Haselmann-ICMLA-2018, Fei-TMM-2020, Jiang-TII-2023, Liu-CVPR-2018, Pirnay-ICIAP-2022, Ristea-Arxiv-2023, Yu-ACMMM-2020, Georgescu-CVPR-2021} are already using the prediction of masked inputs as a surrogate task for anomaly detection. These models form a distinctive subcategory of reconstruction-based methods.
Liu \etal~\cite{Liu-CVPR-2018} proposed a GAN for predicting a future frame based on a few past frames, where anomalies are classified according to the prediction error. Another GAN-based approach \cite{Sabokrou-ACCV-2018} performs joint detection and localization of anomalies via inpainting. The generator of this method learns to inpaint a patch from the input image, while the discriminator learns to identify if the inpainted patch is normal or abnormal. Interestingly, the inpainting task has also been studied in conjunction with vision transformers \cite{Pirnay-ICIAP-2022}.

Generalizing over the method of Liu \etal~\cite{Liu-CVPR-2018}, Yu \etal~\cite{Yu-ACMMM-2020} employed the Cloze task \cite{Luo-Arxiv-2020}, which is about learning to complete the video when certain frames are removed. Georgescu \etal~\cite{Georgescu-CVPR-2021} proposed the masking of the middle box of each temporal cube centered on an object. Anomalies are detected based on the assumption that motion reconstruction for an abnormal object is more difficult than for the normal ones. Fei \etal~\cite{Fei-TMM-2020} proposed the Attribute Restoration Network (ARNet), where attributes such as color and orientation of the input are removed, and the network learns to restore those attributes. The idea is based on the assumption that the anomalous data can be distinguished based on the restoration error. Haselmann \etal~\cite{Haselmann-ICMLA-2018} introduced an approach for surface anomaly detection by erasing a rectangular box from the center of the image and using the interpolation error for the classification of samples into normal or abnormal. Inspired by the success of masked auto-encoders \cite{He-CVPR-2022}, Jiang \etal~\cite{Jiang-TII-2023} proposed a masked Swin Transformer \cite{Liu-ICCV-2021b} that is trained to inpaint masked regions. To cope with the lack of abnormal samples during training, the authors used simulated anomalies. Ristea \etal~\cite{Ristea-Arxiv-2023} employed self-distilled masked auto-encoders to reach an unprecedented level of time efficiency.

Unlike other models based on information masking, we propose a novel approach that incorporates the reconstruction-based functionality into a single neural block, which can be easily integrated into other state-of-the-art anomaly detection models. Our experimental results confirm that our block is a valuable addition to various models, including both CNNs and transformers, which are applied to anomaly detection in a wide range of domains.

\section{Method}

\subsection{Motivation and Overview}

A wide set of computer vision tasks, including anomaly detection \cite{Georgescu-TPAMI-2021,Guo-Arxiv-2021,Li-CVPR-2021,Liu-ICCV-2021,Park-CVPR-2020}, are often addressed with convolutional neural networks (CNNs) \cite{lecun-bottou-ieee-1998,Hinton-NIPS-2012}, due to the impressive performance levels reached by these models, sometimes even surpassing human-level accuracy. The defining component of a CNN architecture is the convolutional layer, which typically comprises multiple filters (kernels) that activate on discriminative local patterns captured within the receptive field of the respective filters. Each filter produces an activation map that is further given as input %, directly or after passing through a pooling operation, 
to the next convolutional layer. Since each filter in the subsequent layer processes all activation maps from the previous layer at once, the local features extracted by the previous layer are combined into more complex features. This sequential processing of features over multiple convolutional layers gives rise to a hierarchy of features during the learning process. Earlier convolutional layers activate on low-level features such as corners or edges, and later layers gradually shift to higher-level features such as car wheels or human body parts, as shown by Zeiler \etal~\cite{Zeiler-ECCV-2014}. Although the learned hierarchy of features is very useful in solving discriminative tasks, CNNs do not have the direct means to model the global arrangement of local features \cite{Sabour-NIPS-2017}, since they do not generalize well to novel viewpoints or affine transformations \cite{Sandru-arXiv-2022}. The inability of grasping the global arrangement of local features is mainly caused by the fact that convolutional filters operate on a limited (and typically small) receptive field, not making use of the context.

% what is the problem solved by masked convolution 
We hereby propose a self-supervised masked convolutional transformer block (SSMCTB), which is aimed at learning to reconstruct masked information based on contextual information. To accurately solve the reconstruction of its masked input, the proposed block is required to employ the context and learn the global structure of the local patterns. Hence, it inherently learns to cope with the problem stated by Sabour \etal~\cite{Sabour-NIPS-2017}, specifically the fact that CNNs lack the proper comprehension of the global arrangement of local features. To embed this learning capability into our block, we structure SSMCTB as a convolutional layer with dilated masked kernels, followed by a transformer module that performs channel attention. We attach a self-supervised loss function to our block in order to minimize the reconstruction error between the masked input and the predicted output. 

We emphasize that SSMCTB is quite flexible, since it can be inserted at any level of almost any CNN or transformer model, generating powerful features that offer the capability of reconstructing masked information based on context. While the ability of learning and harnessing the global arrangement of local patterns is potentially useful in solving a broader set of computer vision tasks, we conjecture that anomaly detection is a natural and immediate application domain for SSMCTB, hence focusing our work in this direction. Indeed, since anomaly detection models are typically trained on normal data only, integrating SSMCTB into a neural model will lead to the learning of features that recover only masked normal data. Hence, when an anomalous sample is given as input during inference, SSMCTB is likely less capable of reconstructing the masked information. This empowers the model to directly estimate the abnormality level of a data sample via the reconstruction error given by SSMCTB. Our claims are supported through the comprehensive set of experiments on image and video anomaly detection presented in Section~\ref{sec_experiments}. 

\begin{figure}[!t]
\begin{center}
\centerline{\includegraphics[width=0.45\linewidth]{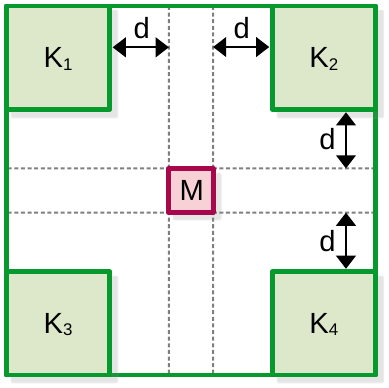}}
\vspace{-0.25cm}
\caption{Our 2D masked convolutional kernel. The visible area of the receptive field is denoted by the regions $\boldsymbol{K}_i, \forall i \in \{1, 2, 3, 4\}$, while the masked area is denoted by $\boldsymbol{M}$. A dilation factor $d$ controls the local or global nature of the visible information with respect to $\boldsymbol{M}$. Best viewed in color.}
\label{fig_kernel}
%\vspace{-0.9cm}
\end{center}
\end{figure}

\subsection{Architecture}

Our initial self-supervised block introduced in \cite{Ristea-CVPR-2022} was formed of a 2D masked convolution and a Squeeze-and-Excitation (SE) module \cite{Hu-CVPR-2018}. To broaden the applicability of our block, we now introduce a 3D masked convolutional layer to replace the 2D masked convolution, whenever this is needed. Moreover, we replace the SE attention module with a modern transformer-based attention module \cite{Dosovitskiy-ICLR-2020,Vaswani-NIPS-2017} to attend to the channels given as output by the masked convolution. We describe the individual components of our block below, while providing a graphical overview of SSMCTB in Figure \ref{fig_sspcab}.

\noindent
\textbf{2D Masked Convolution.} Figure \ref{fig_kernel} shows our 2D masked convolutional kernel, where the corner regions of this kernel (in green color) are the learnable parameters (weights) defining the visible regions of the receptive field. The four learnable sub-kernels are denoted by $\boldsymbol{K}_i \in \mathbb{R}^{k' \times k' \times c}$, $\forall i \in \{1, 2, 3, 4\}$, where the spatial size $k' \in \mathbb{N}^+$ of each sub-kernel is a hyperparameter of our block, while the number of channels $c \in \mathbb{N}^+$ always matches the number of channels of the input tensor. Our masked region $\boldsymbol{M} \in \mathbb{R}^{1 \times 1 \times c}$ (in pink color) is located at the center of the receptive field. Each sub-kernel $\boldsymbol{K}_i$ is located at a configurable distance $d \in \mathbb{N}^+$ (also referred to as  \emph{dilation rate}) from the masked region $\boldsymbol{M}$. To keep the number of hyperparameters to a bare minimum, we fix the spatial size of the masked region to $1 \times 1$. As a result, the spatial size $k$ of the entire receptive field of our 2D masked convolution is $k = 2k' + 2d + 1$. 

Let $\boldsymbol{X} \in \mathbb{R}^{h \times w \times c}$ be the input tensor of the masked convolutional layer, where $c \in \mathbb{N}^+$ denotes the number of channels, and $h, w \in \mathbb{N}^+$ represent the height and width of the input tensor, respectively. When we apply our custom kernel at a given location $(a,b)$ of the input tensor $\boldsymbol{X}$, only the input values that overlap with the sub-kernels $\boldsymbol{K}_i$ are taken into consideration during the masked convolution operation, resulting in a single output value. We underline that our masked convolution is equivalent to convolving the input independently with the sub-kernels $\boldsymbol{K}_i$, where each sub-kernel has a different spatial shift with respect to the current location $(a,b)$, and the resulting values are summed up to produce a single output value. The output value at position $(a,b)$ represents the reconstruction for only one value of the tensor $\boldsymbol{M}$ located at the same position $(a,b)$. To reconstruct the entire tensor $\boldsymbol{M}$, our layer requires the application of $c$ masked convolutional filters, each reconstructing the masked value from a distinct channel at position $(a,b)$. Convolving a single masked filter over the entire input generates a complete activation map. Since there are $c$ masked convolutional filters, the output tensor $\boldsymbol{Z}$ is formed of $c$ activation maps. Our aim is to apply the masked convolution such that every element in the input tensor is masked exactly once, \ie~we want to mask and predict the reconstruction for every spatial location of the input. As such, we set the stride to $1$ and apply a zero-padding of $k'+d$ in each direction. With this configuration in place, the output tensor $\boldsymbol{Z}$ has $h \times w \times c$ components, exactly as the input tensor $\boldsymbol{X}$. To obtain the final values, the output tensor $\boldsymbol{Z}$ is passed through Rectified Linear Units (ReLU) \cite{Nair-ICML-2010}. Finally, we emphasize that $k'$ and $d$ are the only tunable hyperparameters of our masked convolutional layer.

\begin{figure}[!t]
\begin{center}
\centerline{\includegraphics[width=0.68\linewidth]{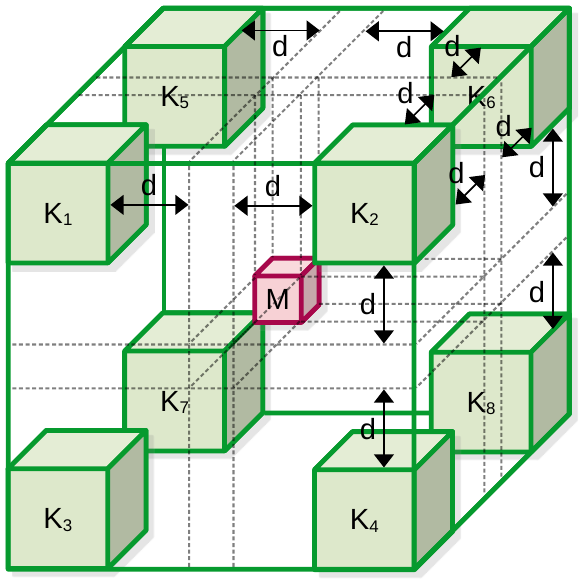}}
\vspace{-0.25cm}
\caption{Our 3D masked convolutional kernel. The visible area of the receptive field is denoted by the regions $\boldsymbol{K}_i, \forall i \in \{1, 2, ..., 8\}$, while the masked area is denoted by $\boldsymbol{M}$. A dilation factor $d$ controls the local or global nature of the visible information with respect to $\boldsymbol{M}$. Best viewed in color.}
\label{fig_kernel_3d}
%\vspace{-0.9cm}
\end{center}
\end{figure}

\noindent
\textbf{3D Masked Convolution.}
Considering that anomaly detection is often applied on 3D inputs, \eg~video or medical scans, some researchers naturally resort to employing 3D CNNs. To this end, we extend our 2D masked convolution to the 3D domain, broadening the applicability of SSMCTB. We thus reformulate the 2D spatial reconstruction task into a more difficult one, which implies learning a global 3D structure of the discovered local patterns. 
Let $\boldsymbol{K}_i \in \mathbb{R}^{k' \times k' \times k' \times c}$, $\forall i \in \{1, 2,..., 8\}$, be the learnable 3D sub-kernels depicted in Figure \ref{fig_kernel_3d}, where $k'$ and $c$ are defined above. The masked region $\boldsymbol{M}$ is located in the center of the 3D kernel, equally distant from the sub-kernels $\boldsymbol{K}_i$. The size of the receptive field of our 3D masked convolution is $k \times k \times k$, where $k=2k' + 2d + 1$.  

To compute the feature response using the 3D masked convolutional layer, the input $\boldsymbol{X} \in \mathbb{R}^{h \times w \times r \times c}$ is convolved with our custom masked kernel, where $r$ represents the depth, and $h$, $w$ and $c$ are defined as before. The 3D filter is applied analogously to the 2D one, the only difference being that the input data and the kernel itself are 3D. The number of 3D convolutional filters is equal to the number of channels $c$, such that the spatial dimension of the output tensor $\boldsymbol{Z} \in \mathbb{R}^{h \times w \times r \times c}$ is identical to that of the input $\boldsymbol{X}$. The 3D masked convolution has the same number of configurable hyperparameters, these being $k'$ and $d$.

\noindent
\textbf{Channel-wise transformer block.}
To better exploit the interdependencies between the different activation maps produced by the masked convolutional layer, we replace the Squeeze-and-Excitation module in SSPCAB \cite{Ristea-CVPR-2022} with a self-attention transformer-based module. The new attention module is able to capture more complex channel-wise interrelations through its higher modeling capacity, as it learns to assign attention weights to the reconstructed information corresponding to each masked convolutional filter in order to reduce the reconstruction error of SSMCTB. 

Let $\boldsymbol{Z} \in \mathbb{R}^{h \times w \times c}$ be the output tensor of a 2D masked convolutional layer with $c$ filters. First, we apply a spatial average pooling, obtaining $\boldsymbol{\hat{Z}} \in \mathbb{R}^{h' \times w' \times c}$, where $h' \leq h$ and $w' \leq w$. The average pooling layer is followed by a reshape operation, obtaining a matrix $\boldsymbol{A} \in \mathbb{R}^{c \times n}$, which contains a vector of $n=h' \cdot w'$ components on each row to represent each masked filter. Next, $\boldsymbol{A}$ is fed into a linear projection layer to obtain the tokens $\boldsymbol{T} \in \mathbb{R}^{c \times d_t}$, which are further summed up with the positional embeddings to obtain the final tokens $\boldsymbol{T}^* \in \mathbb{R}^{c \times d_t}$. %, where $d_t > n'$. 

Let $f$ be a multi-head attention layer with $H \in \mathbb{N}^+$ heads, $g$ a multi-layer perceptron, $\mbox{\emph{norm}}$ a normalization layer, and $\boldsymbol{P}, \boldsymbol{R} \in \mathbb{R}^{c \times d_t}$ some auxiliary tensors. The operations performed inside the transformer are formally described as follows:
\begin{equation}\label{eq_att_1}
\boldsymbol{P} = f(\mbox{\emph{norm}}(\boldsymbol{R})) + \boldsymbol{R},
\end{equation}
\begin{equation}\label{eq_att_2}
\boldsymbol{R} = g(\mbox{\emph{norm}}(\boldsymbol{P}) ) + \boldsymbol{P}.
\end{equation}
As illustrated in Figure~\ref{fig_sspcab}, the whole process described in Eq.~\eqref{eq_att_1} and Eq.~\eqref{eq_att_2} is repeated $L$ times, where $L \in \mathbb{N}^+$ represents the number of transformer blocks inside the transformer module. For the first transformer block, $\boldsymbol{R}$ is initialized with $\boldsymbol{T}^*$. In Eq.~\eqref{eq_att_1}, the sequence of $c$ tokens $\boldsymbol{R}$ is normalized, fed into the multi-head attention layer and added to itself, obtaining $\boldsymbol{P}$. Further, $\boldsymbol{P}$ is normalized, fed into a multi-layer perceptron and also added to itself, according to Eq.~\eqref{eq_att_2}. 

The transformer is aimed at capturing the interaction among all $c$ tokens by encoding each token in terms of the channel-wise contextual information. This is achieved via the multi-head attention layer $f$. Each head $j \in \{1,2,..., H\}$ comprises three learnable weight matrices denoted as $\boldsymbol{W}^{Q_j} \in \mathbb{R}^{d_t \times d_q}$, $\boldsymbol{W}^{K_j} \in \mathbb{R}^{d_t \times d_k}$ and $\boldsymbol{W}^{V_j} \in \mathbb{R}^{d_t \times d_v}$, where $d_q = d_k$. The weight matrices are multiplied with the input tokens $\boldsymbol{R}$, producing the queries $\boldsymbol{Q}_j$, keys $\boldsymbol{K}_j$ and values $\boldsymbol{V}_{\!j}$. In other words, the input sequence $\boldsymbol{R}$ is projected onto these weight matrices to get $\boldsymbol{Q}_j = \boldsymbol{R}\cdot\boldsymbol{W}^{Q_j}$, $\boldsymbol{K}_j = \boldsymbol{R}\cdot\boldsymbol{W}^{K_j}$ and $\boldsymbol{V}_{\!j} = \boldsymbol{R}\cdot\boldsymbol{W}^{V_j}$, respectively. The output $\boldsymbol{Y}_{\!j} \in \mathbb{R}^{c \times d_v}$ of each self-attention head is given by:
\begin{equation}
    \boldsymbol{Y}_{\!j} = \mbox{\emph{softmax}}\left( \frac{\boldsymbol{Q}_j\cdot\boldsymbol{K}^{\top}_j}{\sqrt{d_q}}\right)\cdot \boldsymbol{V}_{\!j},
\end{equation}
where $\boldsymbol{K}^{\top}_j$ is the transpose of $\boldsymbol{K}_j$. The outputs returned by the self-attention heads are simply summed into $\boldsymbol{Y}$, \ie:
\begin{equation}
\boldsymbol{Y} = \sum_{j=1}^H \boldsymbol{Y}_{\!j}. 
\end{equation}
We can now rewrite Eq.~\eqref{eq_att_1} as follows:
\begin{equation}\label{eq_att_3}
\boldsymbol{P} = \boldsymbol{Y}  + \boldsymbol{R}.
\end{equation}
The output sequence $\boldsymbol{R}$ returned by the final transformer block is averaged along the token dimension, obtaining $\boldsymbol{\hat{R}} \in \mathbb{R}^{c \times 1}$, then fed into a sigmoid layer to generate the final attention weight assigned to each channel. Finally, the resulting attention weights are applied to the tensor $\boldsymbol{Z}$, obtaining the reconstructed output denoted by $\boldsymbol{\hat{X}} \in \mathbb{R}^{h \times w \times c}$, as follows:
\begin{equation}
\boldsymbol{\hat{X}} = \boldsymbol{Z} \otimes \sigma(\boldsymbol{\hat{R}}),
\end{equation}
where $\otimes$ denotes the element-wise multiplication, and $\sigma$ denotes the sigmoid layer. The entire processing performed by the transformer module is analogously applied when the preceding layer is a 3D masked convolution.

\subsection{Self-Supervised Reconstruction Loss}

We devise an integrated reconstruction loss to train the proposed SSMCTB in a self-supervised manner. To better cope with outlier values and reduce the sensitivity of the model to outliers, we define the self-supervised objective as the Huber loss between the reconstructed output $\boldsymbol{\hat{X}}$ and the input $\boldsymbol{X}$, replacing the mean squared error (MSE) used by SSPCAB. The self-supervised objective enables our model to learn reconstructing the masked information at every location where the masked filters are applied. Let $G$ denote the SSMCTB function. With this notation, the self-supervised reconstruction loss of our block can be computed as follows: 
\begin{equation}\label{eq_loss_block}
\begin{split}
    \!\! \mathcal{L}_{\mbox{\scriptsize{SSMCTB}}}(G, \boldsymbol{X}) &\!=\! \left\{\begin{array}{ll} \!\!\!\frac{1}{2}\!\cdot\!\left(G(\boldsymbol{X})\!-\! \boldsymbol{X}\right)^2\!, & \!\!\!\!\!\!\!\!\!\!\!\!\!\!\mbox{if} \; |G(\boldsymbol{X})\!-\! \boldsymbol{X}|\!<\!\delta\!\!\!\! \\
    \!\!\!\delta\!\cdot\!\left(|G(\boldsymbol{X})\!-\! \boldsymbol{X}|\!-\!\frac{\delta}{2} \right), & \!\!\!\!\mbox{otherwise}\end{array} \;\right.\\
    &\!=\! \left\{\begin{array}{ll} \!\!\!\frac{1}{2}\!\cdot\!\left(\boldsymbol{\hat{X}}\!-\! \boldsymbol{X}\right)^2\!, & \!\!\!\!\!\!\!\!\!\!\!\!\!\!\mbox{if} \; |\boldsymbol{\hat{X}}\!-\! \boldsymbol{X}|\!<\!\delta \\
    \!\!\!\delta\!\cdot\!\left(|\boldsymbol{\hat{X}}\!-\! \boldsymbol{X}|\!-\!\frac{\delta}{2} \right), & \!\!\!\mbox{otherwise}\end{array} \;\right. ,
\end{split}
\end{equation}
where $\delta \in \mathbb{R}^+$ is a hyperparameter representing the error threshold that determines when to switch from the squared loss (applied for errors below $\delta$) to the absolute loss (applied for errors higher than or equal to $\delta$).

When integrating SSMCTB into some neural network $F$, we can simply add our loss $\mathcal{L}_{\mbox{\scriptsize{SSMCTB}}}$ to the loss function $\mathcal{L}_F$ of the respective neural model, resulting in a new loss function comprising both terms. Formally, the overall loss can be computed as follows:
\begin{equation}\label{eq_loss_total}
\mathcal{L}_{\mbox{\scriptsize{total}}} = \mathcal{L}_F + \lambda \cdot \mathcal{L}_{\mbox{\scriptsize{SSMCTB}}},
\end{equation}
where $\lambda \in \mathbb{R}^+$ is a hyperparameter deciding the importance of $\mathcal{L}_{\mbox{\scriptsize{SSMCTB}}}$ with respect to $\mathcal{L}_F$. Naturally, the hyperparameter $\lambda$ can vary across neural models or visual tasks.

\section{Experiments and Results}
\label{sec_experiments}

\subsection{Data Sets}

We carry out experiments on five benchmarks from various domains, considering the most popular data set choices, \eg~MVTec AD \cite{Bergmann-CVPR-2019}, BRATS \cite{Menze-TMI-2015}, CUHK Avenue \cite{Lu-ICCV-2013}, ShanghaiTech \cite{Luo-ICCV-2017}, whenever such an option is available for a certain domain. For the thermal video domain, we build our own data set.

\noindent
\textbf{MVTec AD.}
MVTec AD \cite{Bergmann-CVPR-2019} has become a standard data set for benchmarking anomaly detection methods applied in inspecting industrial defects. The data set contains over 5,000 images distributed over 15 different categories of textures (10) and objects (5). It comprises 3,629 defect-free training samples, as well as 1,725 test images with and without defects. 

%% Results from Thermal Data
\begin{table}[t!]
\caption{Rare events in our thermal anomaly detection data set along with the frequency of each event type.}
\vspace{-0.2cm}
\centering 
\setlength\tabcolsep{5.0pt}
\small
\begin{tabular}{| l | c |} 
\hline
{Rare Event Type} & {Frequency} \\
\hline
\hline
Activities in restricted zones & 6  \\
Jumping & 4  \\
Reverse driving & 2  \\
Unexpected activities & 2 \\
Unexpected interactions & 14  \\
Unexpected vehicle & 1 \\
\hline
Total & 29\\
\hline
\end{tabular}
\vspace{0.1cm}
\label{table:rare_events} % is used to refer this table in the text
\end{table}

\noindent
\textbf{BRATS.}
BRATS \cite{Menze-TMI-2015} is a multimodal magnetic resonance imaging (MRI) data set for brain tumor segmentation. It is an intrinsically heterogeneous data set that contains brain tumors of different shape, appearance and histology. The data set comprises manually annotated MRI scans acquired by 19 institutions employing different clinical protocols. %BRATS is manually annotated and the annotations are approved by experts. 
To evaluate anomaly detection models, we introduce a novel split of the data set, such that all training images are lesion-free, \ie~all images with lesions are kept for testing. The training set includes 11,280 slices (125 scans), which leaves 27,745 slices (180 scans) for the test set.

\noindent
\textbf{Avenue.}
CUHK Avenue \cite{Lu-ICCV-2013} is one of the most widely-used data sets for video anomaly detection. It contains 16 videos for training and 21 videos for testing. The training videos comprise only normal events, whereas the test videos contain both normal and abnormal events. The data set contains videos from a single surveillance camera. Avenue contains people-related anomalies such as running, walking in the wrong direction, jumping, dancing, loitering and throwing objects. 

\noindent
\textbf{ShanghaiTech.}
ShanghaiTech \cite{Luo-ICCV-2017} is one of the largest benchmarks for video anomaly detection, comprising 330 training and 107 test videos. As in CUHK Avenue, abnormal instances appear only at test time. The data set includes videos from multiple scenes. Examples of anomalies are related to people, \eg~fighting, jumping and stealing, as well as vehicles, \eg~bikes and cars in pedestrian (forbidden) zones. 

\noindent
\textbf{Thermal Rare Event.} To construct the Thermal Rare Event data set, we sampled one week of videos (330 clips) from the Seasons in Drift (SiD) data set \cite{Nikolov-NIPS-2021}. The SiD data set \cite{Nikolov-NIPS-2021} is an unlabeled thermal surveillance data set captured from a single view over a period of 8 months. The data set captures activities near a harbor front during day and night. Each clip is about 2 minutes long and contains 120 frames, being sampled at 1 frame per second (FPS). Out of the 330 clips, there are 29 clips containing rare (anomalous) events. % such as jumping, dancing, running with strollers, embarking to boat, debarking from boat, and unexpected vehicles. 
We manually annotated these rare events at the frame level. In total, our Thermal Rare Event data set contains 36,120 frames for testing and 3,480 frames for training. The list of rare events in our data set along with their respective frequencies are summarized in Table \ref{table:rare_events}. Examples of rare events from different categories are: activities in restricted zones (people sitting, standing, and running close to the pier), jumping (person jumping, group jumping), unexpected activities (doing yoga, smoking), unexpected interactions (running with stroller, embarking to a boat, debarking from a boat, chasing, dancing), unexpected vehicles (different types of trucks). We release the Thermal Rare Event data set along with our code at: \url{https://github.com/ristea/ssmctb/}.

\subsection{Evaluation Measures}

\noindent
\textbf{Image Anomaly Detection.} 
Following Bergmann \etal~\cite{Bergmann-CVPR-2019}, we carry out the evaluation on MVTec AD and BRATS considering the area under the receiver operating characteristics curve (AUROC) and the average precision (AP). To generate the ROC curve, the true positive rate (TPR) is plotted against the false positive rate (FPR). We evaluate both detection and localization performance levels of anomaly detection methods. In anomaly detection, TPR is the proportion of images correctly classified as abnormal, and FPR is the proportion of normal images wrongly classified as abnormal. For the localization task, TPR denotes the proportion of correctly classified abnormal pixels, while FPR represents the proportion of normal pixels incorrectly classified as abnormal. For the localization task, we obtain anomaly segments by applying a threshold to produce a binary decision for each pixel, as described in \cite{Bergmann-CVPR-2019}. The localization AP is obtained by taking the mean at different threshold levels. 

\noindent
\textbf{Video Anomaly Detection.} As the majority of previous works \cite{Ramachandra-PAMI-2020}, we evaluate the detection performance of video anomaly detection methods using the frame-level area under the curve (AUC). To compute the AUC measure, a video frame is marked as abnormal if at least one pixel is abnormal. Inspired by Georgescu \etal~\cite{Georgescu-TPAMI-2021}, we employ both micro AUC and macro AUC. The micro AUC is computed by first concatenating all frames in all videos into a single video, while the macro AUC represents the average of the AUC scores which are independently computed for each single video in the test set. To evaluate the localization performance, we report the region-based detection criterion (RBDC) and the track-based detection criterion (TBDC) proposed by Ramachandra \etal~\cite{Ramachandra-WACV-2020a}. RBDC considers each detected region, marking it as a \emph{true positive} if the intersection over union (IOU) between the detected and the ground-truth anomalous region is greater than $\alpha$. TBDC marks each tracked region as a \textit{true positive} if the overlap with the ground-truth anomalous track is greater than $\beta$. We set the same values for $\alpha$ and $\beta$ as previous works \cite{Ramachandra-WACV-2020a, Georgescu-TPAMI-2021}, \ie~$\alpha=0.1$ and $\beta=0.1$.

\subsection{Implementation Details}
\label{sec_implement}

We choose ten state-of-the-art approaches \cite{Barbalau-ARXIV-2022, Zavrtanik-ICCV-2021, Schulter-ECCV-2022, Georgescu-TPAMI-2021, Liu-ICCV-2021, Park-CVPR-2020, Liu-CVPR-2018,He-CVPR-2022,Vasu-ICCV-2023,Wang-ICDM-2022} for image and video anomaly detection to serve as underlying models, on top of which we add SSPCAB~\cite{Ristea-CVPR-2022} and SSMCTB (ours). We alternatively integrate SSPCAB and SSMCTB directly into the official implementations of the chosen baselines, while preserving all hyperparameter values, \eg~the number of epochs and the learning rate, as specified in the corresponding papers \cite{Barbalau-ARXIV-2022,Zavrtanik-ICCV-2021, Schulter-ECCV-2022, Georgescu-TPAMI-2021, Liu-ICCV-2021, Park-CVPR-2020, Liu-CVPR-2018, He-CVPR-2022,Vasu-ICCV-2023,Wang-ICDM-2022}. Even so, we are unable to exactly reproduce the original results for two baselines methods, \ie~those of Park \etal~\cite{Park-CVPR-2020} and Liu \etal~\cite{Liu-CVPR-2018}. However, our reproduced quantitative results are still close to the originally reported results. For a fair comparison, we compare the models based on SSPCAB and SSMCTB with the reproduced baselines. Additionally, when we repurpose the approach of Park \etal~\cite{Park-CVPR-2020} from the RGB domain to the thermal domain, we modify some hyperparameters, namely the number of epochs and the mini-batch size. 

Following Ristea \etal~\cite{Ristea-CVPR-2022}, we replace the penultimate convolutional layer with SSMCTB in most underlying models. One exception is the architecture of Georgescu \etal~\cite{Georgescu-CVPR-2021}, where SSPCAB and SSMCTB are integrated into the penultimate convolutional layer of the decoder instead of the final classification network. Another exception is the masked auto-enconder \cite{He-CVPR-2022} based on the ViT backbone, where we place SSPCAB and SSMCTB before the first transformer block.

In our previous work~\cite{Ristea-CVPR-2022}, we conducted a set of preliminary experiments to find an optimal value for the hyperparameter $\lambda$ representing the contribution of our self-supervised loss to the total loss defined in Eq.~\eqref{eq_loss_block}, taking values from $0.1$ to $1$ at an interval of $0.1$. Following our previous work~\cite{Ristea-CVPR-2022}, we keep $\lambda=0.1$ across all data sets. However, for two baselines \cite{Liu-ICCV-2021,Schulter-ECCV-2022}, we notice that the magnitude of our loss is too high with respect to the original losses of the respective models, dominating the optimization. Following our previous work~\cite{Ristea-CVPR-2022}, we decrease $\lambda$ to $0.001$ to reduce the dominant influence of our loss on these two particular models \cite{Liu-ICCV-2021,Schulter-ECCV-2022}.

For the channel-wise transformer, we fix the activation map size after the average pooling layer to $1 \times 1$, the token size $d_t$ to $64$, the number of heads $H$ to $4$, as well as the number of successive transformer blocks $L$ to $2$. We discuss results for other transformer configurations in Section~\ref{sec_ablation}.

\subsection{Preliminary Results}

\begin{table}[!t]
\centering 
\caption{Micro AUC scores (in \%) obtained on the Avenue data set with different hyperparameter configurations, varying the kernel size ($k'$), the dilation rate ($d$), the loss type, and the attention type, while integrating SSMCTB into the method of Park \etal~\cite{Park-CVPR-2020}. The top score is highlighted in bold.}\label{tab_ablation}
\vspace{-0.2cm}
\setlength\tabcolsep{4.0pt}
\small
\begin{tabular}{| c | c | c | c | c | c |} 
\hline
 {{Method}} & $\mathcal{L}_{\mbox{\scriptsize{SSMCTB}}}$ & {$d$} & {$k'$} & {Attention} & Micro AUC  \\
 \hline
 \hline
 {Park \etal~\cite{Park-CVPR-2020}} & -  & -  & -  & - & 82.8 \\
  \hline
  & \multirow{4}{*}[0.0ex]{MAE}  & 0 & 1 &  \multirow{4}{*}[0.0ex]{-} & {83.1}  \\
  {+SSMC}&  & 1 & 1 &  & {83.5}   \\
  {(no attention)}&  & 2 & 1 &  & {84.2}   \\
  &  & 3 & 1 &  & {84.4}   \\
  \cline {2-6}
  \hline
\multirow{29}{*}{+SSMCTB}  & \multirow{4}{*}[0.0ex]{MAE}  & 0 & 1 &  \multirow{4}{*}[0.0ex]{CA} & 83.7  \\
  &  & 1 & 1 &  & 84.9   \\
  &  & 2 & 1 &  & 85.5   \\
  &  & 3 & 1 &  & 85.9   \\
  \cline {2-6}
  & \multirow{4}{*}[0.0ex]{MSE} & 0 & 1 & \multirow{4}{*}[0.0ex]{CA} & 84.9  \\
  &  & 1 & 1 &  & 85.7  \\
  &  & 2 & 1 &  & 85.4  \\
  &  & 3 & 1 &  & {86.4}   \\
  \cline {2-6}
  & \multirow{4}{*}[0.0ex]{{SSIM}} & 0 & 1 & \multirow{4}{*}[0.0ex]{CA}  &  {83.3} \\
  &  & 1 & 1 &  &  {85.5} \\
  &  & 2 & 1 &  & {84.9} \\
  &  & 3 & 1 &  & {83.0}  \\
    \cline {2-6}
    & \multirow{4}{*}[0.0ex]{{Huber}} & 0 & 1 & \multirow{4}{*}[0.0ex]{CA}  & {84.2} \\
  &  & 1 & 1 &  & \textbf{{87.0}} \\
  &  & 2 & 1 &  & {86.5} \\
  &  & 3 & 1 &  &  {86.1} \\
  \cline {2-6}
  & \multirow{4}{*}[0.0ex]{{Huber}} & 0 & 2 & \multirow{4}{*}[0.0ex]{CA} & 84.1  \\
  &  & 1 & 2 &  & 84.9  \\
  &  & 2 & 2 &  & 84.8  \\
  &  & 3 & 2 &  & 86.0  \\
  \cline {2-6}
  & \multirow{4}{*}[0.0ex]{{Huber}} & 0 & 3 & \multirow{4}{*}[0.0ex]{CA} & 84.5 \\
  &  & 1 & 3 &  & 85.0  \\
  &  & 2 & 3 &  & 86.4  \\
  &  & 3 & 3 &  & 84.3  \\
    \cline {2-6}
  & \multirow{4}{*}[0.0ex]{{Huber}} & 0 & 1 & \multirow{4}{*}[0.0ex]{SA} & 86.2  \\
  &  & 1 & 1 &  & 84.7  \\
  &  & 2 & 1 &  & 85.8  \\
  &  & 3 & 1 &  & 80.4  \\
  \cline {2-6}
  & {Huber} & 1 & 1 & CA + SA & 86.2 \\
 %\cline {2-8}
 % & \multirow{2}{*}[0.0ex]{MSE} & 1 & 1 & 4 & CA & - & -  \\
 % &  & 1 & 1 & 16 & CA & - & -  \\
 \hline
\end{tabular}
\vspace{0.1cm}
\end{table}

We conduct a series of preliminary experiments on Avenue to determine the hyperparameters of SSMCTB, namely the dilation rate $d$ and the sub-kernel size $k'$. We perform experiments with $d \in$  $\{0,1,2,3\}$ and $k' \in$ $\{1,2,3\}$. We also consider alternative attention types, namely channel attention (CA), spatial attention (SA) and both channel and spatial attention (CA+SA). Additionally, we alternate between multiple losses to self-supervise our block, such as the mean absolute error (MAE), the mean squared error (MSE), the Huber loss, and the Structured Similarity Index Measure (SSIM) loss. For the Huber loss, we set the hyperparameter $\delta$ to the default value, \ie~$\delta=1$.

\begin{table*}[t]
\centering 
\caption{Detection AUROC and localization AUROC/AP (in \%) of three state-of-the-art methods \cite{Schulter-ECCV-2022,Zavrtanik-ICCV-2021,Vasu-ICCV-2023} on MVTec AD, before and after alternatively adding SSPCAB and SSMCTB. The best result for each model and each performance measure is highlighted in bold.}
\setlength\tabcolsep{2.5pt}
\vspace{-0.2cm}
\small
\begin{tabular}{| c | l | ccc | ccc | ccc | ccc | ccc | ccc | ccc |} 
\hline
 & \multirow{9}{*}{Class} &
\multicolumn{9}{c|}{Detection} &
\multicolumn{12}{c|}{Localization}  \\ 
\cline{3-23}
& & \multicolumn{3}{c|}{\multirow{2}{*}{DRAEM \cite{Zavrtanik-ICCV-2021}}} & \multicolumn{3}{c|}{NSA \cite{Schulter-ECCV-2022}}  & \multicolumn{3}{c|}{FastViT \cite{Vasu-ICCV-2023}}  & \multicolumn{6}{c|}{\multirow{2}{*}{DRAEM \cite{Zavrtanik-ICCV-2021}}} & \multicolumn{3}{c|}{NSA \cite{Schulter-ECCV-2022}} & \multicolumn{3}{c|}{FastViT \cite{Vasu-ICCV-2023}} \\
& & \multicolumn{3}{c|}{} & \multicolumn{3}{c|}{(logistic)}  & \multicolumn{3}{c|}{+ NSA \cite{Schulter-ECCV-2022}}  & \multicolumn{6}{c|}{} & \multicolumn{3}{c|}{(logistic)} & \multicolumn{3}{c|}{+ NSA \cite{Schulter-ECCV-2022}} \\
\cline{3-23}
& & \multicolumn{3}{c|}{AUROC} & \multicolumn{3}{c|}{AUROC} & \multicolumn{3}{c|}{AUROC} & \multicolumn{3}{c|}{AUROC} & \multicolumn{3}{c|}{AP} & \multicolumn{3}{c|}{AUROC} & \multicolumn{3}{c|}{AUROC} \\
\cline{3-23}
&  & \rotatebox{90}{Baseline} & \rotatebox{90}{+SSPCAB~}   & \rotatebox{90}{+SSMCTB~} & \rotatebox{90}{Baseline}  & \rotatebox{90}{+SSPCAB} & \rotatebox{90}{+SSMCTB}  & \rotatebox{90}{Baseline} & \rotatebox{90}{+SSPCAB} & \rotatebox{90}{+SSMCTB} & \rotatebox{90}{Baseline}  &
\rotatebox{90}{+SSPCAB} & \rotatebox{90}{+SSMCTB} & \rotatebox{90}{Baseline} &
\rotatebox{90}{+SSPCAB} & \rotatebox{90}{+SSMCTB} & \rotatebox{90}{Baseline} &
\rotatebox{90}{+SSPCAB} & \rotatebox{90}{+SSMCTB} & \rotatebox{90}{Baseline} &
\rotatebox{90}{+SSPCAB} & \rotatebox{90}{+SSMCTB} 
\\
%\cline{3-11}
%&  & AUROC   & AUROC  & AUROC  & AP  & AP & AP & AUROC  & AUROC  & AUROC \\
\hline
\hline
\multirow{5}{*}[0.0ex]{\rotatebox{90}{Texture}} & Carpet &   97.0  &  \textbf{98.2} & 96.8 &  95.6 &  \textbf{97.5}    & 96.1 & 95.5 & 95.6 & \textbf{95.7} &   95.5 & 95.0 & \textbf{95.8} & 53.5 & \textbf{59.4} & 55.2 & 95.5 & \textbf{97.5} & 95.6  & 95.4 & 95.5 & \textbf{95.6} \\

& Grid        & 99.9 & \textbf{100}   & \textbf{100} & 99.9   &  99.9    & \textbf{100} & 99.8 & 99.8 & \textbf{100} & \textbf{99.7} &  99.5 & \textbf{99.7} & 65.7 & 61.1  & \textbf{69.7} & \textbf{99.2} & \textbf{99.2} & \textbf{99.2} & 99.1 & 99.2 & \textbf{99.4}  \\

& Leather     &  \textbf{100}  &  \textbf{100} & \textbf{100} & 99.9    &  99.9  & \textbf{100} & 99.9 & \textbf{100} & \textbf{100} &  98.6 & \textbf{99.5} & 97.6  & 75.3 & \textbf{76.0} & 65.5  & 99.5 & 99.5 & \textbf{99.6} & 99.1 & 99.2 & \textbf{99.3} \\

& Tile        &  99.6  & \textbf{100}  & \textbf{100}  & \textbf{100}      & \textbf{100}     & \textbf{100} & 99.9 & \textbf{100} & \textbf{100} &  99.2 & \textbf{99.3} & \textbf{99.3} & 92.3 & 95.0 & \textbf{95.7} & \textbf{99.3} & 99.2 & 99.1 & 99.3 & \textbf{99.5} & 99.4 \\

& Wood        &  99.1  &  99.5 & \textbf{100} & 97.5     &  97.7     & \textbf{97.8} & 97.5 & 97.7 & \textbf{97.8} & 96.4 & \textbf{96.8} & 94.8 & \textbf{77.7} & 77.1 & 75.6 & 90.7 & 90.4 & \textbf{93.5} & 90.5 & 90.8 & \textbf{91.0} \\
\hline

\multirow{10}{*}[0.0ex]{\rotatebox{90}{Object}} & Bottle    &  99.2 &  98.4 & \textbf{99.4} &  \textbf{97.7}    &   \textbf{97.7}    & \textbf{97.7} & 97.6 & 97.5 & \textbf{97.7} &   99.1  & 98.8 & \textbf{99.2} & 86.5 & 87.9 & \textbf{89.9} & 98.3 & 98.3 & \textbf{98.4} & \textbf{98.3} & 98.2 & \textbf{98.3} \\

& Cable       &  91.8  & \textbf{96.9}  & 94.1 & 94.5     &  95.6     &  \textbf{96.1} & 94.3 & \textbf{94.7} & 94.6 & 94.7 & \textbf{96.0} & 95.5 & 52.4 & 57.2 & \textbf{61.6} & 96.0 & 96.6 & \textbf{97.5} & 95.8 & 96.1 & \textbf{96.3} \\

& Capsule     &  98.5  &  \textbf{99.3} & 97.1 &  95.2    &  95.4    & \textbf{95.5} & 95.2 & 95.2 & \textbf{95.4} & \textbf{94.3} & 93.1 & 93.4 & 49.4  & 50.2 & \textbf{52.0} & 97.6 & 97.2 & \textbf{97.9} & 97.5 & \textbf{97.6} & {97.5} \\

& Hazelnut    & \textbf{100}   & \textbf{100}  & \textbf{100} & 94.7      &  94.2     & \textbf{97.1} & 94.6 & \textbf{94.8} & 94.6 & 99.7 & \textbf{99.8} & 99.5 & \textbf{92.9} & 92.6 & 89.1 & 97.6 & \textbf{97.9} & \textbf{97.9} & \textbf{97.5} & {97.4} & {97.4} \\

& Metal Nut    &  98.7  &  \textbf{100} & \textbf{100} & 98.7      &  99.0     & \textbf{99.5} & 98.5 & \textbf{98.6} & \textbf{98.6} & \textbf{99.5} & 98.9 & 99.3 & 96.3 & \textbf{98.1} & 94.7 & 98.4 & \textbf{98.6} & 98.3 & \textbf{98.1} & 98.0 & \textbf{98.1} \\

& Pill         &  98.9  &  \textbf{99.8} & \textbf{98.8} & 99.2     &   99.2    & \textbf{99.5} & 99.2 & 99.3 & \textbf{99.4} &  \textbf{97.6} & 97.5 & 97.4 & 48.5 & \textbf{52.4} & 46.9 & 98.5 & \textbf{98.8} & 98.4 & \textbf{98.4} & 98.3 & \textbf{98.4} \\

& Screw        &  93.9  &  97.9 & \textbf{99.0} & 90.2      &  \textbf{91.1}     & 90.4 & 90.2 & 90.3 & \textbf{90.5} & 97.6 & \textbf{99.8} & 99.5 & 58.2 & \textbf{72.0} & 70.1 & \textbf{96.5} & 96.2 & 96.4 & 96.4 & 96.6 & \textbf{96.7} \\

& Toothbrush   &  \textbf{100}  & \textbf{100}  & \textbf{100} & \textbf{100}      &  \textbf{100}     & \textbf{100} & \textbf{100} & \textbf{100} & \textbf{100} & 98.1 & 98.1 & \textbf{99.0} & 44.7 & 51.0 & \textbf{69.0} & 94.9 & 95.3 & \textbf{95.4} & 94.0 & \textbf{94.2} & \textbf{94.2} \\

& Transistor   &  93.1  & 92.9  & \textbf{96.0} & 95.1     &   95.6    & \textbf{96.2} & {95.1} & \textbf{95.2} & {95.1} &  \textbf{90.9} & 87.0 & 89.1 & \textbf{50.7} & 48.0 & 45.8 & 88.0 & 87.1 & \textbf{88.3} & 88.1 & 88.2 & \textbf{88.3} \\

& Zipper       &  \textbf{100}  &  \textbf{100} & \textbf{100} & 99.8 &   99.8    & \textbf{99.9} & \textbf{99.8} & \textbf{99.8} & \textbf{99.8} & 98.8 & \textbf{99.0} & \textbf{99.0} & \textbf{81.5} & 77.1 & 76.5 & 94.2 & 94.5 & \textbf{94.7} & {93.2} & \textbf{93.3} & \textbf{93.3} \\
\hline
& Overall      &  98.0  &  \textbf{98.9} & 98.7 & 97.2      &   97.5   &  \textbf{97.7} & 97.1 & 97.2 & \textbf{97.3} & \textbf{97.3} & 97.2 & 97.2 & 68.4 & 70.3  & \textbf{70.5} & 96.3 & 96.4 & \textbf{96.7} & 96.0 & 96.1 & \textbf{96.2} \\
\hline
\end{tabular}
\vspace{-0.2cm}
\label{table:ImageAnomaly} % is used to refer this table in the text
\end{table*}

%% Results from Medical Data
\begin{table}[t!]
\centering 
\caption{Detection AUROC and localization AUROC/AP (in \%) of three state-of-the-art methods \cite{Schulter-ECCV-2022,Zavrtanik-ICCV-2021,Vasu-ICCV-2023} on BRATS, before and after alternatively adding SSPCAB and SSMCTB. Additional results obtained by converting DRAEM to use 3D convolutions and integrating the 3D SSMCTB are also reported. The best result for each model and each performance measure is highlighted in bold. The top score for each metric is shown in red.}
\vspace{-0.2cm}
\setlength\tabcolsep{1.8pt}
\small
\begin{tabular}{| l | c | c | c |} 
\hline
 \multirow{7}{*}{Method} & \multicolumn{2}{c|}{AUROC} & \multirow{2}{*}{\rotatebox{90}{~Localization AP~}} \\
\cline{2-3}
   &  \rotatebox{90}{~Detection} & \rotatebox{90}{~Localization~~} &  \\
 \hline % inserts single horizontal line
 \hline
NSA \cite{Schulter-ECCV-2022} & 53.66  & 74.90 & 61.09   \\
NSA + SSPCAB \cite{Ristea-CVPR-2022} & 54.91  & 75.30 & 62.37   \\
NSA + SSMCTB (Ours) & \textbf{60.09} & \textcolor{red}{\textbf{77.09}} & \textbf{64.55}    \\
\hline
FastViT \cite{Vasu-ICCV-2023} + NSA \cite{Schulter-ECCV-2022} & 56.71  & 75.52 & 64.17   \\
FastViT \cite{Vasu-ICCV-2023} + NSA \cite{Schulter-ECCV-2022} + SSPCAB \cite{Ristea-CVPR-2022} & 56.99  & 76.41 & 64.24   \\
FastViT \cite{Vasu-ICCV-2023} + NSA \cite{Schulter-ECCV-2022} + SSMCTB (Ours) & \textcolor{red}{\textbf{60.65}} & \textcolor{red}{\textbf{77.09}} & \textcolor{red}{\textbf{65.48}}    \\
\hline
DRAEM \cite{Zavrtanik-ICCV-2021} & 41.06 & 42.40 & 45.41 \\
DRAEM + SSPCAB \cite{Ristea-CVPR-2022} & 44.19  & 46.66 & 46.89  \\
DRAEM + SSMCTB (Ours)  & \textbf{50.27}  & \textbf{53.98} & \textbf{50.75}  \\
\hline
3D DRAEM \cite{Zavrtanik-ICCV-2021} & 43.74 & 44.12 & 45.97 \\
%3D DRAEM + 3D SSMCTB (Ours)  &  \textbf{51.01} & \textbf{56.22} & \textbf{51.98}  \\
{3D DRAEM + 3D SSMCTB (Ours)}  &  \textbf{53.70} & \textbf{58.47} & \textbf{52.79}  \\
\hline
\end{tabular}
\vspace{-0.2cm}
\label{table:MedicalAnomaly} % is used to refer this table in the text
\end{table}

%python Train.py --method pred 1>train_out.log 2>train_err.log &
We employ the method of Park \etal~\cite{Park-CVPR-2020} in our preliminary experiments, since this is the most lightweight and unpretentious method among the chosen ones \cite{Barbalau-ARXIV-2022,Zavrtanik-ICCV-2021, Schulter-ECCV-2022, Georgescu-TPAMI-2021, Liu-ICCV-2021, Park-CVPR-2020, Liu-CVPR-2018,He-CVPR-2022,Vasu-ICCV-2023,Wang-ICDM-2022}. The corresponding micro AUC scores are presented in Table~\ref{tab_ablation}. Except for a single SSMCTB configuration based on spatial attention (SA), all other SSMCTB configurations bring performance improvements over the approach of Park \etal~\cite{Park-CVPR-2020} (first row). Our first set of preliminary experiments is aimed at evaluating the capacity of the standalone masked convolution. Even without the attention module, our masked convolution brings gains higher than $1\%$ for $d=2$ and $d=3$. While adding the attention module is definitely useful, we conclude that it is clearly not the only factor responsible for the reported performance gains. To compare the losses on the one hand, and attention types on the other, we fix $k'=1$. When alternating between MAE, MSE, SSIM and Huber as our self-supervised loss, we generally observe higher performance with Huber loss. We thus continue the experiments with Huber loss. Regarding the attention type, we note that channel attention (CA) generally leads to better results than spatial attention (SA). Hence, for the remaining experiments, we employ the transformer module based on channel attention. We continue by increasing the size of the sub-kernels, without obtaining further performance gains. We obtain the best micro AUC ($86.7\%$) with $d=1$ and $k'=1$, while using channel attention. We make another attempt to further boost the performance by combining the channel and spatial attention (CA+SA), while fixing $d=1$ and $k'=1$. This attempt is also unsuccessful. Our final SSMCTB configuration, which we employ across all underlying models and data sets, is based on $d=1$, $k'=1$ and channel attention.

We underline that the corresponding hyperparameters for SSPCAB were tuned in a similar manner, in our previous work \cite{Ristea-CVPR-2022}. Hence, we simply use the already tuned hyperparameters for SSPCAB. Importantly, we underline that our observations above are mostly consistent with those reported in our previous work~\cite{Ristea-CVPR-2022}, \ie~both SSPCAB and SSMCTB use channel attention, a dilation rate of $d=1$ and sub-kernels of size $k'=1$. The only difference is that SSMCTB is based on the Huber loss instead of the MSE loss. We should also emphasize that it is not common for anomaly detection data sets to have validation splits. Since the training set contains normal instances only, keeping a representative training subset (with both normal and abnormal examples) for validation is not possible. This is the reason behind our decision to avoid hyperparameter tuning for each model and data set. We believe that this evaluation procedure is more fair because it avoids overfitting in hyperparameter space.

\begin{figure}[t!]
\begin{center}
\centerline{\includegraphics[width=0.9\linewidth]{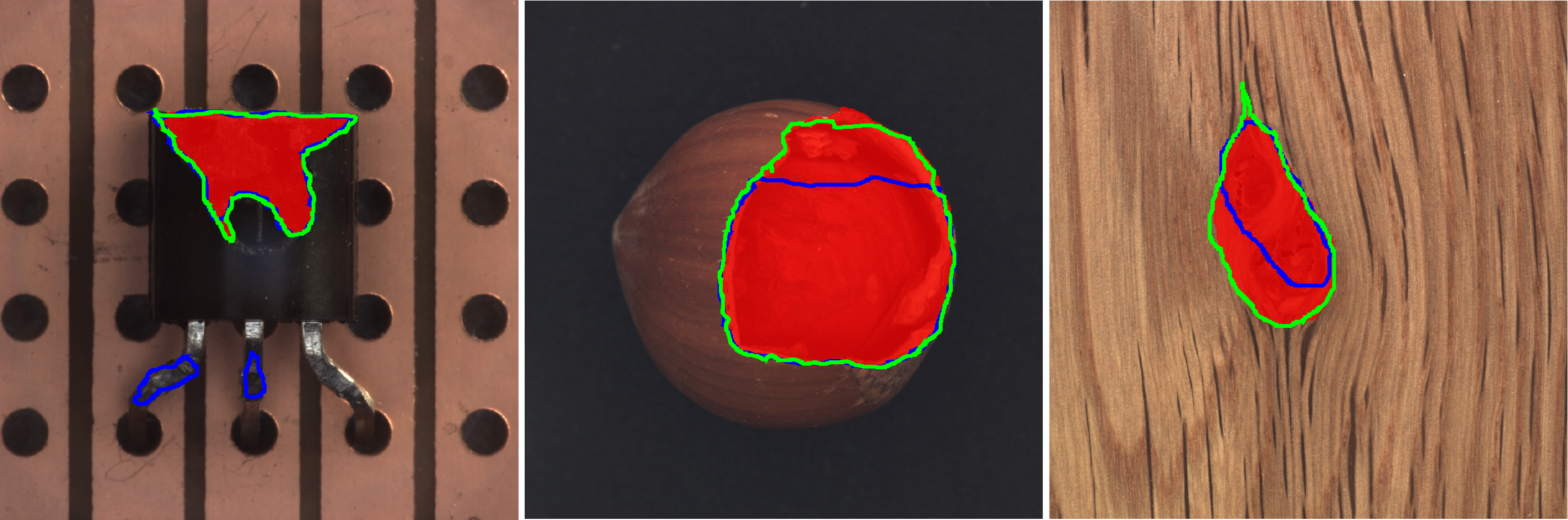}}
\vspace{-0.2cm}
\caption{Examples of image-level anomaly localization results from MVTec AD given by DRAEM \cite{Zavrtanik-ICCV-2021}, before (blue contour) and after (green contour) integrating SSMCTB. The ground-truth anomalies are shown in red. Best viewed in color.}
\label{fig_results_MVtec}
\vspace{-0.2cm}
\end{center}
\end{figure}

\subsection{Anomaly Detection in Images}

\noindent
\textbf{Baselines.}
% describe baselines
We introduce SSMCTB into three state-of-the-art baselines for image anomaly detection on MVTec AD, namely a self-supervised model based on natural synthetic anomalies (NSA) \cite{Schulter-ECCV-2022}, a discriminatively trained reconstruction anomaly embedding model (DRAEM) \cite{Zavrtanik-ICCV-2021}, and a version of FastViT \cite{Vasu-ICCV-2023} based on the T8 backbone. Since FastViT \cite{Vasu-ICCV-2023} is not particularly designed for anomaly detection, we actually couple it with NSA \cite{Schulter-ECCV-2022} to perform the designated task. All three baselines are based on very recent studies, attaining strong results on MVTec AD. 
The NSA approach of Sh\"{u}lter \etal~\cite{Schulter-ECCV-2022} generates synthetic anomalies using Poisson image editing, blending scaled patches of different sizes from separate images. In this way, it generates a wide range of synthetic anomalies that are similar to natural irregularities. 
DRAEM \cite{Zavrtanik-ICCV-2021} comprises a reconstructive network and a discriminative network to detect and localize anomalies. The reconstructive network is based on a simple auto-encoder architecture which learns to reconstruct original images from artificially corrupted images. The discriminative network is a U-Net that learns to segment the introduced artifacts (corrupted regions).

\noindent
\textbf{Results on MVTec AD.}
We report the results on MVTec AD in Table \ref{table:ImageAnomaly}. Considering the detection results, we observe that adding SSPCAB and SSMCTB leads to superior results for all three models. Considering the localization results, the AUROC scores of DRAEM do not show any improvements when adding SSPCAB and SSMCTB. However, the localization AP of DRAEM exhibits gains of around $2\%$ by adding SSPCAB and SSMCTB. In addition, the localization AUROC values of NSA and FastViT+NSA grow when SSPCAB and SSMCTB are introduced into the respective architectures.

In Figure~\ref{fig_results_MVtec}, we present some examples of qualitative results from MVTec AD, obtained by DRAEM \cite{Zavrtanik-ICCV-2021}, before and after adding SSMCTB. In all shown cases, we observe that the anomaly localization results are better aligned with the ground-truth regions when SSMCTB is integrated into DRAEM.

\begin{figure}[t!]
\begin{center}
\centerline{\includegraphics[width=1.0\linewidth]{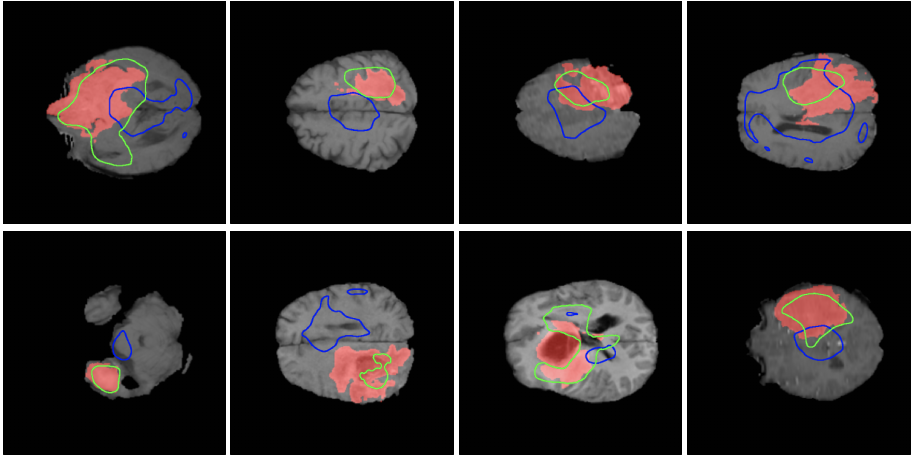}}
\vspace{-0.2cm}
\caption{Examples of image-level anomaly localization results from BRATS given by DRAEM \cite{Zavrtanik-ICCV-2021}, before (blue contour) and after (green contour) integrating SSMCTB. The ground-truth anomalies are shown in red. Best viewed in color.}
\label{fig_results_BraTS}
\vspace{-0.4cm}
\end{center}
\end{figure}

\noindent
\textbf{Results on BRATS.} In Table \ref{table:MedicalAnomaly}, we present the brain lesion detection and localization results obtained by the anomaly detection models \cite{Zavrtanik-ICCV-2021,Schulter-ECCV-2022,Vasu-ICCV-2023} on BRATS, before and after adding SSPCAB and SSMCTB, respectively. Remarkably, we notice that the results of DRAEM, NSA and FastViT+NSA show significant performance improvements when integrating SSMCTB. Moreover, the performance gain brought by SSMCTB is always higher than the gain brought by SSPCAB. When taking advantage of the 3D nature of the MRI scans by employing the 3D SSMCTB, we attain even higher performance with DRAEM. 

In Figure~\ref{fig_results_BraTS}, we present several examples of qualitative results from BRATS, given by DRAEM \cite{Zavrtanik-ICCV-2021}, before and after adding SSMCTB. In general, the localization results based on SSMCTB exhibit a higher overlap with the ground-truth regions, explaining why SSMCTB leads to superior performance levels.

\subsection{Anomaly Detection in Videos}

\begin{table*}[t]
\centering 
\caption{Micro-averaged frame-level AUC, macro-averaged frame-level AUC, RBDC, and TBDC scores (in \%) of various state-of-the-art methods on Avenue and ShanghaiTech. Among the existing models, we select seven models \cite{Barbalau-ARXIV-2022,Georgescu-TPAMI-2021,Liu-ICCV-2021,Liu-CVPR-2018,Park-CVPR-2020,He-CVPR-2022,Wang-ICDM-2022} to show results before and after including SSPCAB and SSMCTB, respectively. The best result for each underlying model is highlighted in bold. The top score for each metric is shown in red.}
\vspace{-0.2cm}
\setlength\tabcolsep{5.0pt}
\small
\begin{tabular}{| l | c | c | c | c | c | c | c | c |  c | c |} 
\hline
 \multirow{3}{*}{Method} & \multicolumn{4}{c|}{Avenue} & \multicolumn{4}{c|}{ShanghaiTech}  \\
 \cline{2-9}
 & \multicolumn{2}{c|}{AUC} & \multirow{2}{*}{RBDC} & \multirow{2}{*}{TBDC} & \multicolumn{2}{c|}{AUC} & \multirow{2}{*}{RBDC} & \multirow{2}{*}{TBDC} \\
 \cline{2-3}
 \cline{6-7}
 & Micro & Macro &  &  & Micro & Macro &  &  \\
 \hline % inserts single horizontal line
 \hline
Liu \etal~\cite{Liu-BMVC-2018} & 84.4 & - & - & - & - & - & - & - \\
Sultani \etal~\cite{Sultani-CVPR-2018} & - & - & - & - & - & 76.5 & - & -  \\
 \cline{2-3}
 \cline{6-7}
% Lee \etal~\cite{Lee-ICASSP-2018} & \multicolumn{2}{c|}{87.2} & - & - & \multicolumn{2}{c|}{76.2} & - & -  \\
%  \hline
Ionescu \etal~\cite{Ionescu-WACV-2019} & 88.9 & - & - & - & - & - & - & -  \\
Nguyen \etal~\cite{Nguyen-ICCV-2019} & 86.9 & - & - & - & - & - & - & - \\
Ionescu \etal~\cite{Ionescu-CVPR-2019} & 87.4 & 90.4 & 15.77 & 27.01 & 78.7 & 84.9 & 20.65 & 44.54  \\
 \cline{2-3}
Wu \etal~\cite{Wu-TNNLS-2019} & \multicolumn{2}{c|}{86.6} & - & - & - & - & - & -  \\
Lee \etal~\cite{Lee-TIP-2019} & \multicolumn{2}{c|}{90.0} & - & - & - & - & - & - \\
%  \hline
\cline{2-3}
Yu \etal~\cite{Yu-ACMMM-2020} & 89.6 & - & - & - & 74.8 & - & - & -  \\
 \cline{2-3}
Ramachandra \etal~\cite{Ramachandra-WACV-2020a} & \multicolumn{2}{c|}{72.0} & 35.80 & 80.90 & - & - & - & -  \\
Ramachandra \etal~\cite{Ramachandra-WACV-2020b} & \multicolumn{2}{c|}{87.2} & 41.20 & 78.60 & - & - & - & -  \\
 \cline{6-7}
Tang \etal~\cite{Tang-PRL-2020} & \multicolumn{2}{c|}{85.1} & - & - & \multicolumn{2}{c|}{73.0} & - & -  \\
Dong \etal~\cite{Dong-Access-2020} & \multicolumn{2}{c|}{84.9} & - & - & \multicolumn{2}{c|}{73.7} & - & - \\
Doshi \etal~\cite{Doshi-CVPRW-2020a}  & \multicolumn{2}{c|}{86.4} & - & - & \multicolumn{2}{c|}{71.6} & - & - \\
Sun \etal~\cite{Sun-ACMMM-2020}  & \multicolumn{2}{c|}{89.6} & - & - & \multicolumn{2}{c|}{74.7} & - & -  \\
Wang \etal~\cite{Wang-ACMMM-2020} & \multicolumn{2}{c|}{87.0} & - & - & \multicolumn{2}{c|}{79.3} & - & -  \\
\cline{2-3}
 \cline{6-7}
%  \hline
Astrid \etal~\cite{Astrid-ICCVW-2021} & 84.7 & - & - & - & 73.7 & - & - & - \\
Astrid \etal~\cite{Astrid-BMVC-2021} & 87.1 & - & - & - & 75.9 & - & - & - \\
Georgescu \etal~\cite{Georgescu-CVPR-2021} &  91.5 & 92.8 & 57.00 & 58.30 & 82.4 & 90.2 & 42.80 & 83.90  \\

 \hline
Liu \etal~\cite{Liu-CVPR-2018} & 85.1 & 81.7 & 19.59 & 56.01 & 72.8 & 80.6 & 17.03 & 54.23 \\
Liu \etal~\cite{Liu-CVPR-2018} + SSPCAB \cite{Ristea-CVPR-2022} & 87.3 & 84.5 & 20.13 & 62.30 & 74.5 & 82.9 & 18.51 & 60.22 \\
%Liu \etal~\cite{Liu-CVPR-2018} + SSMCTB (Ours) & 89.1 & \textbf{84.8} & 20.11 & 63.27 & \textbf{74.6} & 83.3 & 18.22 & \textbf{62.65} \\
Liu \etal~\cite{Liu-CVPR-2018} + SSMCTB (Ours) & \textbf{89.5} & \textbf{84.6} & \textbf{23.79} & \textbf{66.03} & \textbf{74.6} & \textbf{83.9} & \textbf{19.13} & \textbf{61.65} \\

\hline
{He} \etal~\cite{He-CVPR-2022} & 84.0 & 85.6 & - & - & 74.3 & 81.1 & - & - \\ 
{He} \etal~\cite{He-CVPR-2022} + SSPCAB \cite{Ristea-CVPR-2022} & 85.1 & 85.8 & - & - & 74.5 & \textbf{81.9} & - & - \\ 
{He} \etal~\cite{He-CVPR-2022} + SSMCTB (Ours) & \textbf{86.4} & \textbf{86.5} & - & - & \textbf{76.1} & 81.6 & - & - \\ 
\hline
Park \etal~\cite{Park-CVPR-2020} & 82.8 & 86.8 & - & - & 68.3 & 79.7 & - & - \\ 
Park \etal~\cite{Park-CVPR-2020} + SSPCAB \cite{Ristea-CVPR-2022} & 84.8 & \textbf{88.6} & - & - & 69.8 & 80.2 & - & -  \\ 
Park \etal~\cite{Park-CVPR-2020} + SSMCTB (Ours) & \textbf{87.0} & 87.7 & - & - & \textbf{70.6} & \textbf{80.3} & - & - \\ 
 \hline
Liu \etal~\cite{Liu-ICCV-2021} & 89.9 & 93.5 & 41.05 & 86.18 & 74.2 & 83.2 & 44.41  & 83.86 \\ 
Liu \etal~\cite{Liu-ICCV-2021} + SSPCAB \cite{Ristea-CVPR-2022} &  \textbf{90.9} & 92.2 & \textbf{62.27} & \textcolor{red}{\textbf{89.28}} &  \textbf{75.5} & 83.7  & 45.45 & 84.50 \\ 
Liu \etal~\cite{Liu-ICCV-2021}  + SSMCTB (Ours) &  89.6 & \textcolor{red}{\textbf{93.9}} & 46.49 & 86.43 &  75.2 & \textbf{83.8}  & \textbf{45.86} & \textbf{84.69} \\

 \hline
Georgescu \etal~\cite{Georgescu-TPAMI-2021} &  92.3 & 90.4 & 65.05 & \textbf{66.85} &  82.7 & 89.3 & \textbf{41.34} & 78.79 \\
Georgescu \etal~\cite{Georgescu-TPAMI-2021} + SSPCAB \cite{Ristea-CVPR-2022} & 92.9 & \textbf{91.9} & 65.99 & 64.91 & \textbf{83.6} & \textbf{89.5} &  40.55 & \textbf{83.46}  \\
Georgescu \etal~\cite{Georgescu-TPAMI-2021} + SSMCTB (Ours) & \textcolor{red}{\textbf{93.2}} & 91.8 & \textcolor{red}{\textbf{66.04}} & 65.12 & 83.3 & \textbf{89.5} &  40.52 & 81.93 \\
\hline
Wang \etal~\cite{Wang-ICDM-2022} & 89.8 & 91.6 & 28.40 & 82.40 & 77.8 & 80.2 & 44.42	& 81.73 \\
Wang \etal~\cite{Wang-ICDM-2022} + SSPCAB~\cite{Ristea-CVPR-2022} & 90.5 & 91.5 & 28.68 & 83.13 & \textbf{77.9} & 80.4 & 45.12 & 77.04 \\
Wang \etal~\cite{Wang-ICDM-2022} + SSMCTB (Ours) & \textbf{91.1} & \textbf{92.2} & \textbf{28.73} & \textbf{83.23} & \textbf{77.9} & \textbf{81.2} & \textbf{45.72} & \textbf{83.18} \\
\hline
% \hline
B\u{a}rb\u{a}l\u{a}u \etal~\cite{Barbalau-ARXIV-2022} & 91.6 & \textbf{92.5} & 47.83 & 85.26 & \textcolor{red}{\textbf{83.8}} & 90.5 & 47.14 & 85.61 \\
B\u{a}rb\u{a}l\u{a}u \etal~\cite{Barbalau-ARXIV-2022} + 3D SSMCTB (Ours) & 91.6 & 92.4 & \textbf{49.01} & \textbf{85.94} & 83.7 & \textcolor{red}{\textbf{90.6}} & \textcolor{red}{\textbf{47.73}} & \textcolor{red}{\textbf{85.68}} \\
\hline
\end{tabular}
\vspace{-0.2cm}
\label{table:VideoAnomaly} % is used to refer this table in the text
\end{table*}

\begin{figure}[t]
\begin{center}
\centerline{\includegraphics[width=1.0\linewidth]{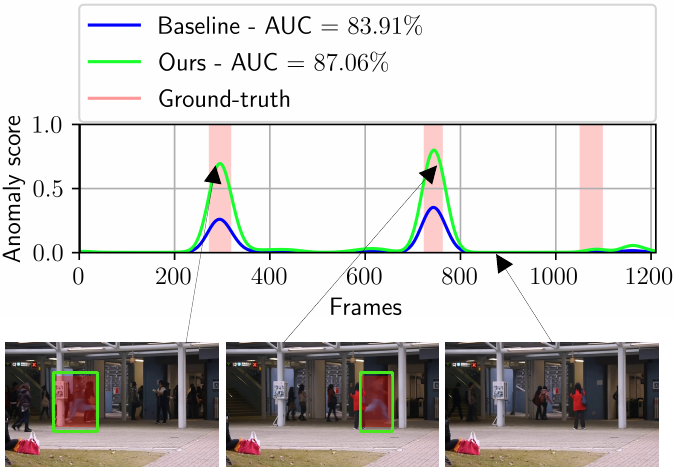}}
\vspace{-0.2cm}
\caption{Frame-level anomaly scores of the method of Georgescu \etal~\cite{Georgescu-TPAMI-2021}, before (baseline) and after (ours) integrating SSMCTB, for test video 02 from the Avenue data set. Anomaly localization results correspond to the model based on SSMCTB. Best viewed in color.}
\label{fig_results_avenue}
\vspace{-0.2cm}
\end{center}
\end{figure}

\begin{figure}[t]
\begin{center}
\centerline{\includegraphics[width=1.0\linewidth]{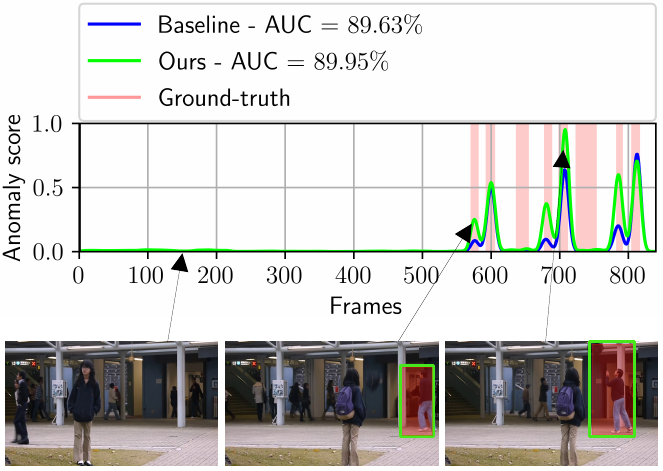}}
\vspace{-0.2cm}
\caption{Frame-level anomaly scores of the method of Georgescu \etal~\cite{Georgescu-TPAMI-2021}, before (baseline) and after (ours) integrating SSMCTB, for test video 10 from the Avenue data set. Anomaly localization results correspond to the model based on SSMCTB. Best viewed in color.}
\label{fig_results_avenue_2}
\vspace{-0.2cm}
\end{center}
\end{figure}

\begin{figure}[t]
\begin{center}
\centerline{\includegraphics[width=1.0\linewidth]{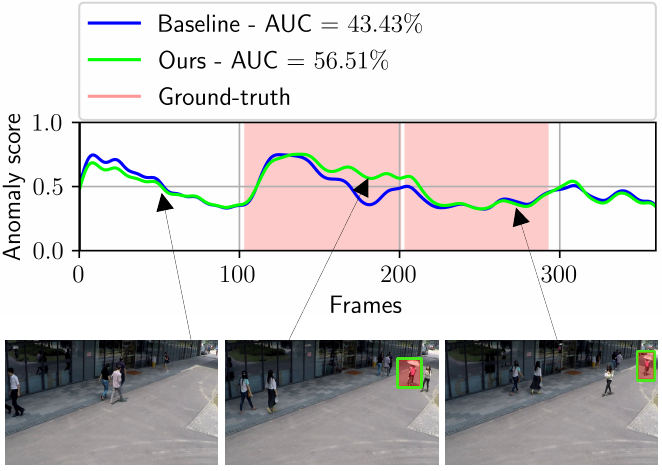}}
\vspace{-0.2cm}
\caption{Frame-level anomaly scores of the method of Liu 
\etal~\cite{Liu-ICCV-2021}, before (baseline) and after (ours) integrating SSMCTB, for test video 02\_0164 from the ShanghaiTech data set. Anomaly localization results correspond to the model based on SSMCTB. Best viewed in color.}
\label{fig_results_shtech}
\vspace{-0.2cm}
\end{center}
\end{figure}

\noindent
\textbf{Baselines.}
% describe baselines
We select six recent methods \cite{Liu-CVPR-2018,Park-CVPR-2020,Georgescu-TPAMI-2021,Liu-ICCV-2021, Barbalau-ARXIV-2022,Wang-ICDM-2022} yielding state-of-the-art performance on Avenue and ShanghaiTech. 
Liu \etal~\cite{Liu-CVPR-2018} proposed a GAN-based framework to detect anomalies based on the future frame prediction error. 
Park \etal~\cite{Park-CVPR-2020} presented a memory-based auto-encoder classifying anomalies based on the reconstruction error. The model comprises a memory module that memorizes prototypes of normal samples. 
Liu \etal~\cite{Liu-ICCV-2021} employed a hybrid framework based on flow reconstruction and frame prediction, using the accumulated error to detect anomalies. 
Georgescu \etal~\cite{Georgescu-TPAMI-2021} introduced a training scheme where the latent subspaces of appearance and motion auto-encoders are improved by performing gradient ascent on pseudo-anomalies during training. Wang \etal~\cite{Wang-ICDM-2022} proposed a novel transformer-based spatio-temporal auto-encoder for
object-centric video anomaly detection, which employs an input perturbation approach to improve the reconstruction capability of the model. B\u{a}rb\u{a}l\u{a}u \etal~\cite{Barbalau-ARXIV-2022} extended the previous work of Georgescu \etal~\cite{Georgescu-CVPR-2021} with two 3D transformer-based self-supervised multi-task architectures trained on new sets of proxy tasks. Among the two versions proposed in \cite{Barbalau-ARXIV-2022}, we opt for SSMTL++v2. We included this 3D model \cite{Barbalau-ARXIV-2022} because it serves as a good baseline for applying our 3D SSMCTB. 

We also experiment with the recently proposed masked auto-encoder framework \cite{He-CVPR-2022}, which is based on the ViT backbone \cite{Dosovitskiy-ICLR-2020}. We add this seventh baseline model to further demonstrate the applicability of SSMCTB to vision transformers.

\noindent
\textbf{Results on RGB videos.}
We present the results on Avenue and ShanghaiTech in Table~\ref{table:VideoAnomaly}. As for the image anomaly detection experiments, we compare the results of the underlying models before and after adding SSPCAB \cite{Ristea-CVPR-2022} and SSMCTB, respectively. For the method of Liu \etal~\cite{Liu-CVPR-2018}, both SSPCAB and SSMCTB lead to performance improvements, but the gains brought by SSMCTB are always higher than those brought by SSPCAB. Since the methods of He \etal~\cite{He-CVPR-2022} and Park \etal~\cite{Park-CVPR-2020} are only capable of detecting anomalies at the frame level, we only report their frame-level micro and macro AUC scores. The vanilla masked auto-encoder obtains competitive results on both Avenue and ShanghaiTech. On Avenue, SSMCTB brings higher gains to the masked auto-encoder than SSPCAB. On ShanghaiTech, SSMCTB is better than SSPCAB in terms of the micro AUC, but SSPCAB exhibits higher macro AUC gains. In summary, both SSMCTB and SSPCAB improve the masked auto-encoder, with SSMCTB having the upper hand. Considering the results of Park \etal~\cite{Park-CVPR-2020} on Avenue, SSMCTB leads to higher gains in terms of the micro AUC (from $82.8\%$ to $87.0\%$), while SSPCAB leads to a higher macro AUC (from $86.8\%$ to $88.6\%$). On ShanghaiTech, we observe higher gains after adding SSMCTB rather than SSPCAB. Moving on to the object-centric models of Liu \etal~\cite{Liu-ICCV-2021} and Georgescu \etal~\cite{Georgescu-TPAMI-2021}, we observe that the top gains are mainly shared between SSPCAB and SSMCTB. However, for the object-centric transformer of Wang \etal~\cite{Wang-ICDM-2022}, SSMCTB yields higher gains than SSPCAB.

%% Main Results from Thermal Data 
\begin{table}[t]
\centering 
\caption{Micro and macro AUC scores (in \%) on Thermal Rare Event, obtained while alternatively including SSPCAB \cite{Ristea-CVPR-2022} and SSMCTB into the method of Park \etal~\cite{Park-CVPR-2020}.}
\vspace{-0.2cm}
\setlength\tabcolsep{5.0pt}
\small
\begin{tabular}{| l | c | c |} 
\hline
\multirow{2}{*}{Method} & \multicolumn{2}{c|}{AUC} \\
\cline{2-3}
  &  Micro & Macro \\
\hline % inserts single horizontal line
\hline
Park \etal~\cite{Park-CVPR-2020} &  53.2 & 66.5  \\
Park \etal~\cite{Park-CVPR-2020} + SSPCAB & 53.6  & \textbf{66.6}  \\
Park \etal~\cite{Park-CVPR-2020} + SSMCTB (Ours) & \textbf{58.9}  & \textbf{66.6} \\
\hline
\end{tabular}
\vspace{-0.2cm}
\label{table:ResultsThermalData} % is used to refer this table in the text
\end{table}

When integrating our 3D SSMCTB into the 3D architecture presented in \cite{Barbalau-ARXIV-2022}, we observe performance improvements according to most metrics. Overall, SSMCTB leads to the highest performance levels on Avenue for three metrics, namely the micro AUC ($93.2\%$), the macro AUC ($93.9\%$) and the RBDC ($66.04\%$). At the same time, SSPCAB attains the highest TBDC score ($89.28\%$) on Avenue. On ShanghaiTech, it appears that the best scores are obtained by adding the 3D SSMCTB into the underlying model of B\u{a}rb\u{a}l\u{a}u \etal~\cite{Barbalau-ARXIV-2022}, since our 3D SSMCTB brings performance gains for three metrics.

In Figures \ref{fig_results_avenue} and \ref{fig_results_avenue_2}, we illustrate the anomaly detection performance on two test videos from Avenue, before and after integrating SSMCTB into the model of Georgescu \etal~\cite{Georgescu-TPAMI-2021}. Our approach produces superior frame-level anomaly scores, being able to detect the person running in the first video (Figure \ref{fig_results_avenue}) and the person throwing an object in the second one (Figure \ref{fig_results_avenue_2}). Moreover, in the second video, we also notice that SSMCTB increases the anomaly score for the penultimate abnormal event, resolving the false negative detection of the baseline. Similarly, in Figure \ref{fig_results_shtech}, we show the effect of adding SSMCTB into the architecture of Liu \etal~\cite{Liu-ICCV-2021} applied on a test video from ShanghaiTech. Once again, SSMCTB improves the frame-level detection performance, being able to detect the person riding a bike in a pedestrian area, which is forbidden. SSMCTB correctly raises the anomaly scores for about 50 video frames, starting at around frame index 150, thus reducing the false negative rate. 

\begin{figure}[t]
\begin{center}
\centerline{\includegraphics[width=1.0\linewidth]{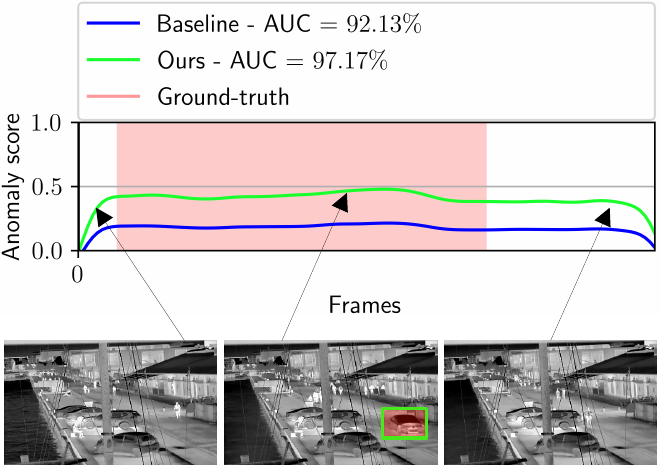}}
\vspace{-0.2cm}
\caption{Frame-level anomaly scores of the method of Park \etal~\cite{Park-CVPR-2020}, before (baseline) and after (ours) integrating SSMCTB, for test video 39 from the Thermal Rare Event data set. Anomaly localization results correspond to the model based on SSMCTB. Best viewed in color.}
\label{fig_results_thermal}
\vspace{-0.2cm}
\end{center}
\end{figure}

\noindent
\textbf{Results on thermal videos.}
Since texture is not present in the thermal domain, there is no need to apply very deep architectures, as noticed by Nikolov \etal~\cite{Nikolov-NIPS-2021}. Moreover, object detectors pre-trained on natural images do not work equally well in the thermal domain due to the distribution shift. To this end, the object-centric \cite{Barbalau-ARXIV-2022,Georgescu-TPAMI-2021,Liu-ICCV-2021,Wang-ICDM-2022} and very deep \cite{Liu-CVPR-2018} baselines attain very poor results (micro AUC values under $50\%$). Hence, we resort to employing the architecture of Park \etal~\cite{Park-CVPR-2020} as underlying model for SSPCAB and SSMCTB. As shown in Table \ref{table:ResultsThermalData}, the chosen baseline attains a micro AUC of $53.2\%$ and a macro AUC of $66.5\%$. Both SSPCAB and SSMCTB seem to have a positive influence on the micro AUC score, but the gains of the latter block are significantly higher (above $5\%$). In summary, the results reported on Thermal Rare Event demonstrate the utility of SSMCTB, further confirming the gains observed on RGB video data sets.

In Figure~\ref{fig_results_thermal}, we show the anomaly detection performance on a test video from Thermal Rare Event, before and after integrating SSMCTB into the model of Park \etal~\cite{Park-CVPR-2020}. SSMCTB leads to important gains in terms of the frame-level scores, being able to detect the vehicle moving backwards.

\begin{table}[t]
\centering
\caption{Inference time (in milliseconds) per example for three frameworks \cite{Liu-CVPR-2018,Georgescu-TPAMI-2021,Zavrtanik-ICCV-2021}, before and after integrating SSPCAB and SSMCTB, respectively. The running times are measured on an Nvidia GeForce GTX 3090 GPU with 24 GB of VRAM.}
\label{tab_inference_time}
\noindent
\small
\begin{tabular}{|l|cccc|}
\hline
\multirow{7}{*}{Method} & \multicolumn{4}{c|}{Time (ms)} \\
\cline{2-5}
& \rotatebox{90}{Baseline} & \rotatebox{90}{+SSPCAB} & \rotatebox{90}{+SSMCTB} & \rotatebox{90}{{+3D SSMCTB}\;}\\
\hline
\hline
{Liu \etal~\cite{Liu-CVPR-2018}} & 2.1 & 2.4 & 2.5 & - \\ 
\hline
{Georgescu \etal~\cite{Georgescu-TPAMI-2021}} & 1.5 & 1.7 & 1.8 & - \\
\hline
{{Zavrtanik} \etal~\cite{Zavrtanik-ICCV-2021}} & 26.4 & - & - & 26.6 \\
\hline
\end{tabular}
\end{table}

% Ablation by applying SSPCAB to different stages with different dilation rates
\begin{table}[t]
\centering 
\caption{Micro AUC (in \%) on Avenue by incorporating SSMCTB into different conv blocks of the decoder proposed by Park \etal~\cite{Park-CVPR-2020}. Along with the block placement, we also vary the dilation rate $d$.}
\vspace{-0.2cm}
\setlength\tabcolsep{2.5pt}
\small
\begin{tabular}{| l | c | c | c | } 
\hline
Method & Decoder Conv Block & $d$ & {Micro AUC} \\
 \hline % inserts single horizontal line
 \hline
{Park \etal~\cite{Park-CVPR-2020}} & - & - & 82.8 \\
\hline
\multirow{16}{*}{+SSMCTB} & early & 0 & 83.6 \\
& early & 1 & 83.2 \\
& early & 2 & 84.6  \\
& early & 3 & 84.7  \\
& early & 4 & 84.9  \\
\cline{2-4}
& middle & 0 & 84.3  \\
& middle & 1 & 83.2  \\
& middle & 2 & 84.5  \\
& middle & 3 & 85.9  \\
& middle & 4 & 85.7  \\
\cline{2-4}
& late & 0 & 86.2  \\
& late & 1 & \textbf{87.0}  \\
& late & 2 & 85.9  \\
& late & 3 & 84.6   \\
& late & 4 & 85.8   \\
\cline{2-4}
& all & 4,3,1 & 85.1 \\
\hline 
\end{tabular}
\vspace{-0.2cm}
\label{table:ablation1_avenue} % is used to refer this table in the text
\end{table}

\subsection{Inference Time}

Regardless of the underlying framework \cite{Barbalau-ARXIV-2022, Zavrtanik-ICCV-2021, Schulter-ECCV-2022, Georgescu-TPAMI-2021, Liu-ICCV-2021, Park-CVPR-2020, Liu-CVPR-2018,He-CVPR-2022,Vasu-ICCV-2023,Wang-ICDM-2022}, similar to Ristea \etal~\cite{Ristea-CVPR-2022}, we add only one instance of SSMCTB, usually replacing the penultimate convolutional layer. Considering that the channel attention from SSPCAB is replaced with a channel-wise transformer block in SSMCTB, we might expect a slightly higher processing time. To assess the amount of extra time added by SSMCTB, we present the running times before and after integrating SSPCAB and SSMCTB into two state-of-the-art frameworks \cite{Liu-CVPR-2018,Georgescu-TPAMI-2021} in Table~\ref{tab_inference_time}. For both baseline models, the time added by SSMCTB is at most $0.1$ ms higher than the time taken by SSPCAB. Moreover, the computational time of SSMCTB does not exceed a difference of $0.4$ ms with respect to the original baselines.
Another important question is how does the 3D version of SSMCTB impact the running time. To answer this question, we take the DRAEM model \cite{Zavrtanik-ICCV-2021} and measure the running time before and after adding the 3D SSMCTB. The reported time measurements show that the running time increase due to the 3D SSMCTB is still marginal, being around $0.2$ ms. Hence, the processing delays caused by the introduction of the 2D or 3D SSMCTB versions are within the same range. In summary, we consider that the accuracy gains brought by SSMCTB outweigh the marginal running time expansions reported in Table~\ref{tab_inference_time}.

\subsection{Ablation Study}
\label{sec_ablation}

\noindent
\textbf{Block placement.}
Across all the experiments presented so far, recall that we introduce a single SSMCTB, which is usually placed near the end of the architecture (penultimate convolutional layer), as mentioned in Section~\ref{sec_implement}. The number of blocks as well as their placement should be tuned on some validation set, which could lead to higher performance gains. However, anomaly detection data sets do not commonly contain a validation set and there is no way to keep a number of training samples for validation, as the training set comprises only normal examples. To this end, we employed a single configuration (one block, closer to the output) to fairly demonstrate the universality of SSMCTB. Certainly, this choice might not always be optimal. Hence, we perform ablation experiments by incorporating SSMCTB at different decoder levels of the network proposed by Park \etal~\cite{Park-CVPR-2020}, considering different dilation rates ($d$). We vary the dilation rate along with the block placement, because Du\c{t}\u{a} \etal~\cite{Duta-ICCVW-2021} observed that higher dilation rates are suitable for earlier dilated convolutional layers, and lower dilation rates are suitable for dilated convolutional layers closer to the output.

In Table \ref{table:ablation1_avenue}, we show the corresponding results on the Avenue data set. We start by adding SSMCTB into the earliest stage of the decoder (first conv block), progressively moving the block to the layers closer to the output of the decoder, until we reach the very last one. For each decoder level (early, middle, late), we vary the dilation rate to find a suitable value. We attain the best micro AUC ($87.0\%$) when integrating SSMCTB into the last conv block of the decoder, while using a dilation rate of $d=1$. A dilation rate of $d=4$ seems suitable when placing SSMCTB at an earlier stage, while, for the middle stage placement, the optimal dilation rate appears to be $d=3$. Interestingly, these results are consistent with the observation made by Du\c{t}\u{a} \etal~\cite{Duta-ICCVW-2021}, although their observation applies to dilated convolutions, while ours applies to masked convolutions. Nevertheless, all the results are consistently better than the baseline ($82.8\%$), regardless of the block placement or the dilation rate. We do not observe major improvements when integrating multiple blocks, concluding that integrating a single SSMCTB is sufficient.

\begin{table}[t!]
\centering 
\caption{Micro AUC (in \%) on Avenue by incorporating SSMCTB into the model of Park \etal~\cite{Park-CVPR-2020}, while varying the size of the masked region $\boldsymbol{M}$.}
\vspace{-0.2cm}
\setlength\tabcolsep{5.0pt}
\small
\begin{tabular}{| l | c | c |} 
\hline
 Method & Size of $\boldsymbol{M}$ & Micro AUC \\
 \hline % inserts single horizontal line
 \hline
 {Park \etal~\cite{Park-CVPR-2020}} 
 & - & $82.8$ \\
\hline
 \multirow{3}{*}{+SSMCTB} & $1 \times 1 $ & $87.0$ \\
 & $2 \times 2 $ & $85.6$  \\
 & $3 \times 3 $ & $84.9$  \\

\hline
\end{tabular}
\vspace{-0.2cm}
\label{table:ablation2_avenue} % is used to refer this table in the text
\end{table}

\begin{table}[t!]
\centering 
\caption{Micro AUC (in \%) on Avenue by incorporating SSMCTB into the model of Park \etal~\cite{Park-CVPR-2020}, while varying the hyperparameters of the channel-wise transformer, namely the activation map size ($h' \times w'$) after the average pooling layer, the token size ($d_t$) after the projection layer, the number of heads ($H$), as well as the number of successive transformer blocks ($L$).}
\vspace{-0.2cm}
\setlength\tabcolsep{5.0pt}
\small
\begin{tabular}{| l | c | c | c | c | c |} 
\hline
{Method} & $h' \times w'$ & $d_t$ & {$H$} & $L$ &  {Micro AUC} \\
\hline % inserts single horizontal line
\hline
{Park \etal~\cite{Park-CVPR-2020}}
& - & - & - & - & 82.8 \\
\hline
\multirow{16}{*}{+SSMCTB} & $1 \times 1 $ & 64 & 4 & 2 & {87.0} \\
& $2 \times 2 $ & 64 & 4 & 2 & {85.3} \\
& $3 \times 3 $ & 64 & 4 & 2 & {85.2} \\
& $4 \times 4 $ & 64 & 4 & 2 & {85.6} \\
\cline{2-6}
& $1 \times 1 $ & 16 & 4 & 2 & {84.6}  \\
& $1 \times 1 $ & 32 & 4 & 2 & {85.6}  \\
& $1 \times 1 $ & 64 & 4 & 2 & {87.0} \\
& $1 \times 1 $ & 128 & 4 & 2 & {85.1} \\
\cline{2-6}
& $1 \times 1 $ & 64 & 3 & 2 & {85.6}  \\
& $1 \times 1 $ & 64 & 4 & 2 & {87.0} \\
& $1 \times 1 $ & 64 & 5 & 2 & {87.0} \\
& $1 \times 1 $ & 64 & 6 & 2 & {84.8} \\
\cline{2-6}
& $1 \times 1 $ & 64 & 4 & 1 & 85.1  \\
& $1 \times 1 $ & 64 & 4 & 2 & {87.0} \\
& $1 \times 1 $ & 64 & 4 & 3 & 84.0 \\
\hline
\end{tabular}
\vspace{-0.2cm}
\label{table:ablation3_avenue} % is used to refer this table in the text
\end{table}

\noindent
\textbf{Size of masked region.}
Increasing the size of the masked region $\boldsymbol{M}$ can lead to a harder reconstruction task, at each location where our masked convolution is applied. However, it is unclear if making the task harder leads to better results. To this end, we vary the spatial size of $\boldsymbol{M}$, considering three options: $1\times 1$, $2\times 2$ and $3\times 3$. We present the corresponding results in Table \ref{table:ablation2_avenue}. The empirical results indicate that increasing the size of $\boldsymbol{M}$ leads to lower anomaly detection scores. Hence, we conclude that a size of $1\times 1$ for the masked region $\boldsymbol{M}$ is optimal.

\noindent
\textbf{Transformer architecture.}
In Table \ref{table:ablation3_avenue}, we present further ablation experiments for the channel-wise transformer module. We keep the underlying model of Park \etal~\cite{Park-CVPR-2020} and report the results on the Avenue data set. As variations for the transformer module, we consider the following hyperparameters: the activation map size ($h' \times w'$) after the average pooling layer, the token size ($d_t$) after the projection layer, the number of heads ($H$), as well as the number of successive transformer blocks ($L$).

First, we analyze how activation maps of different dimensions, given as output by the average pooling layer placed right before the transformer, influence the results. We observe that shrinking the maps to $1 \times 1$ gives the best micro AUC ($87.0\%$). The optimal configuration of the average pooling layer (producing activation maps of $1 \times 1$) is equivalent to global average pooling. For the projection layer, we consider output dimensions in the set $d_t \in \{16,32,64,128\}$. The optimal size for the projection layer is $d_t=64$. We consider transformer modules having $3$ to $6$ heads. The empirical evidence indicates that using $H=4$ or $H=5$ heads leads to equally good results. Finally, we experiment with transformer modules having $1$ to $3$ blocks. The best performance is achieved with $L=2$ successive transformer blocks. We underline that all transformer configurations surpass the baseline model \cite{Park-CVPR-2020}.

\begin{table}[!t]
\centering 
\caption{Micro AUC (in \%) on Avenue by incorporating SSMCTB into the model of Park \etal~\cite{Park-CVPR-2020}, while varying the hyperparameter $\delta$ of the Huber loss.}
\label{tab_huber_ablation}
\vspace{-0.2cm}
\setlength\tabcolsep{4.0pt}
\small
\begin{tabular}{| c | c | c |} 
\hline
 {{Method}} & $\delta$ &  Micro AUC  \\
 \hline
 \hline
  {Park \etal~\cite{Park-CVPR-2020}} & - & 82.8 \\
  \hline
  \multirow{3}{*}{+SSMCTB}  & 0.5 & 84.1 \\
  & 1 & 87.0 \\
  & 2 &  85.8 \\
  \hline
  \end{tabular}
\vspace{0.1cm}
\end{table}

\begin{table}[!t]
\centering 
\caption{Micro AUC (in \%) on Avenue by incorporating SSMCTB into the model of Park \etal~\cite{Park-CVPR-2020}, while switching between dilated and masked convolution. Different values for the dilation rate $d$ are tested for the two operations.}
\label{tab_comparison_with_dilation}
\vspace{-0.2cm}
\setlength\tabcolsep{4.0pt}
\small
\begin{tabular}{| c | c | c |} 
\hline
 {{Method}} & $d$ &  Micro AUC  \\
 \hline
 \hline
  {Park \etal~\cite{Park-CVPR-2020}} & - & 82.8 \\
  \hline
  \multirow{3}{*}{+SSDCTB (dilated conv)}  & 1 & 85.1 \\
  & 2 & 83.3  \\
  & 3 & 85.0  \\
  \hline
  \multirow{3}{*}{+SSMCTB (masked conv)}  & 1 & 87.0  \\
    & 2 & 85.5  \\
    & 3 & 85.9  \\
  \hline
  \end{tabular}
\vspace{0.1cm}
\end{table}

\noindent
\textbf{Huber loss hyperparameter.}
Huber loss is the combination of the $L_1$ (MAE) and $L_2$ (MSE) losses (see Eq.~\eqref{eq_loss_block}), where $\delta$ is a hyperparameter representing the threshold that switches between the two loss functions. To study the effect of $\delta$, we consider different values for the hyperparameter $\delta \in \{0.5, 1, 2 \}$, reporting the results in Table \ref{tab_huber_ablation}. We find that the maximum improvement corresponds to $\delta=1$, but the other values of $\delta$ also lead to superior results compared to the baseline.

\noindent
\textbf{Comparison with dilated convolution.}
In Table \ref{tab_comparison_with_dilation}, we compare the dilated convolution against the proposed masked convolution, alternating between the two operations inside SSMCTB. We denote the block based on dilated convolution through the acronym SSDCTB. When comparing the two convolutional operations, we consider multiple dilation rates between $1$ and $3$. The experiments show that the proposed masked convolution outperforms the dilated convolution, regardless of the dilation rate. This confirms that the two operations are not equivalent, essentially revealing the importance of the self-supervised task based on reconstructing the masked region $\boldsymbol{M}$ situated in the center of the receptive field. 

\section{Conclusion}

In this paper, we extended our previous work \cite{Ristea-CVPR-2022} by introducing SSMCTB, a novel neural block composed of a masked convolutional layer and a channel-wise transformer module, which predicts a masked region in the center of the convolutional receptive field. Our neural block is trained in a self-supervised manner, via a reconstruction loss of its own.
To show the benefits of using SSMCTB in anomaly detection, we integrated our block into a series of image and video anomaly detection methods \cite{Barbalau-ARXIV-2022, Zavrtanik-ICCV-2021, Schulter-ECCV-2022, Georgescu-TPAMI-2021, Liu-ICCV-2021, Park-CVPR-2020, Liu-CVPR-2018, He-CVPR-2022,Vasu-ICCV-2023,Wang-ICDM-2022}. In addition, we included two new benchmarks from domains that were not previously considered in our previous work \cite{Ristea-CVPR-2022}, namely medical images and thermal videos. Moreover, we extended the 2D masked convolution to a 3D masked convolution, broadening the applicability of the self-supervised block to 3D neural architectures. To showcase the utility of the new 3D SSMCTB, we integrated our 3D block into two 3D networks (3D DRAEM and SSMTL++v2) for anomaly detection in image and video, respectively.
Our empirical results across multiple benchmarks and underlying models indicate that SSMCTB brings performance improvements in a vast majority of cases. Furthermore, with the help of SSMCTB, we are able to obtain new state-of-the-art levels on the widely-used Avenue and ShanghaiTech data sets. We consider this as a major achievement, which would not have been possible without SSMCTB.

In future work, we aim to apply our novel self-supervised block on other tasks, aside from anomaly detection. For example, due to the self-supervised loss computed with respect to the masked region, our block could be integrated into various neural architectures to perform self-supervised pre-training, before applying the respective models to downstream tasks. Interestingly, the pre-training could be performed at multiple architectural levels, \ie~wherever the block is added into the model.

% use section* for acknowledgment
\ifCLASSOPTIONcompsoc
  % The Computer Society usually uses the plural form
  \section*{Acknowledgments}
\else
  % regular IEEE prefers the singular form
  \section*{Acknowledgment}
\fi

This work was supported by a grant of the Romanian Ministry of Education and Research, CNCS - UEFISCDI, project no. PN-III-P2-2.1-PED-2021-0195, contract no. 690/2022, within PNCDI III. The research leading to these results has also received funding from the NO Grants 2014-2021, under project ELO-Hyp contract no. 24/2020. Additionally, this work has been funded by Milestone Systems through the Milestone Research Programme at AAU, and by SecurifAI. % Moreover, this article has benefited from the support of the Romanian Young Academy, which is funded by Stiftung Mercator and the Alexander von Humboldt Foundation for the period 2020-2022. The work is also supported by starting grant (GR010) and VR starting grant (2016-05543).

%\newpage
%%%%%%%%% REFERENCES
{\small
\bibliographystyle{ieeetr}
\bibliography{references}
}

\begin{IEEEbiography}[{\includegraphics[width=0.6in,height=0.78in,clip,keepaspectratio]{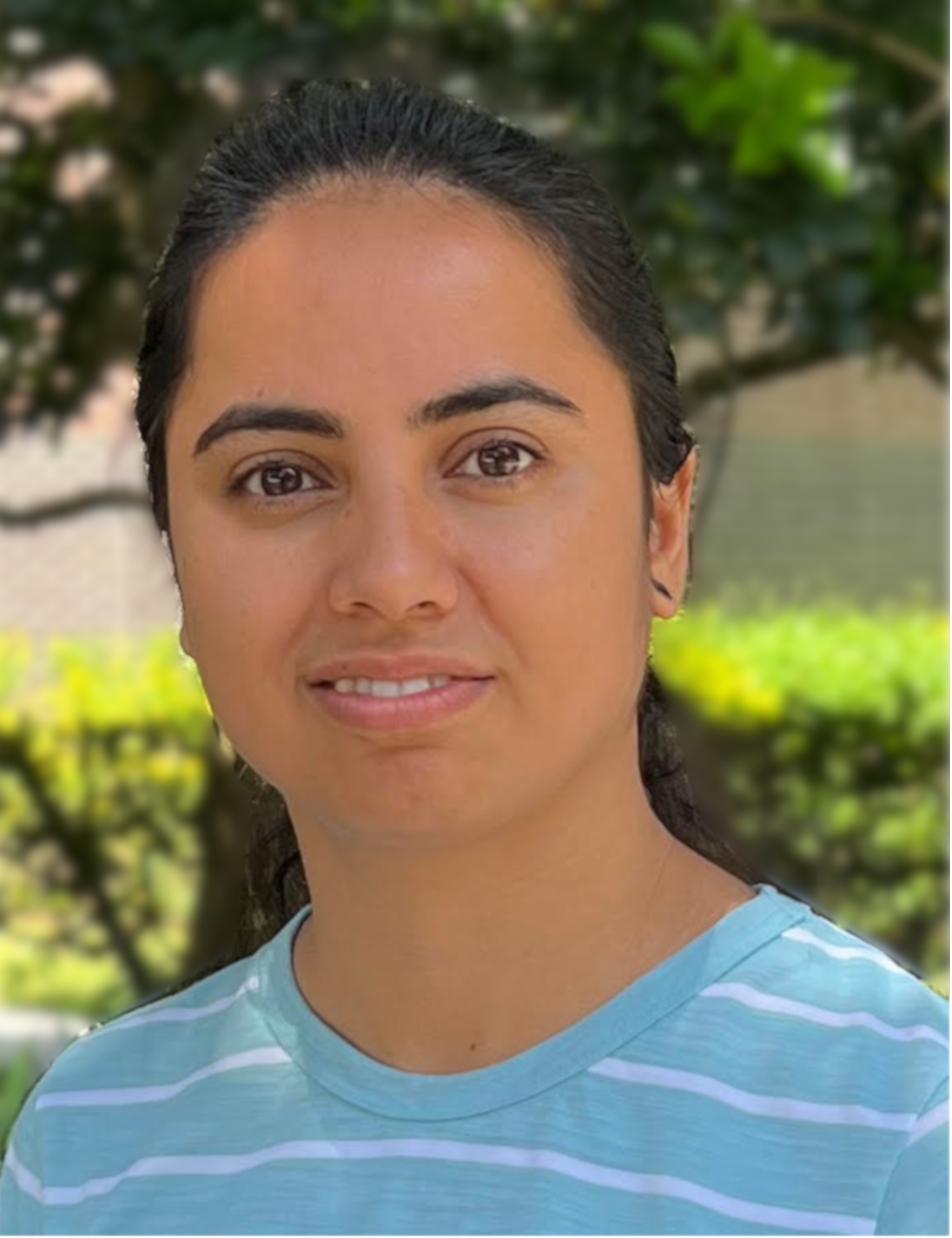}}]
{Neelu Madan} is a Ph.D. student at the Department of Media Technology at Aalborg University, Denmark. She received her M.Sc.~degree in Computer Science with a major as Interactive Systems and Visualization from the University of Duisburg-Essen, Germany. Her research interests include artificial intelligence, computer vision, machine learning, and deep learning. She is the author of a paper accepted for oral presentation at CVPR 2022.
\end{IEEEbiography}

\begin{IEEEbiography}[{\includegraphics[width=0.6in,height=0.78in,clip,keepaspectratio]{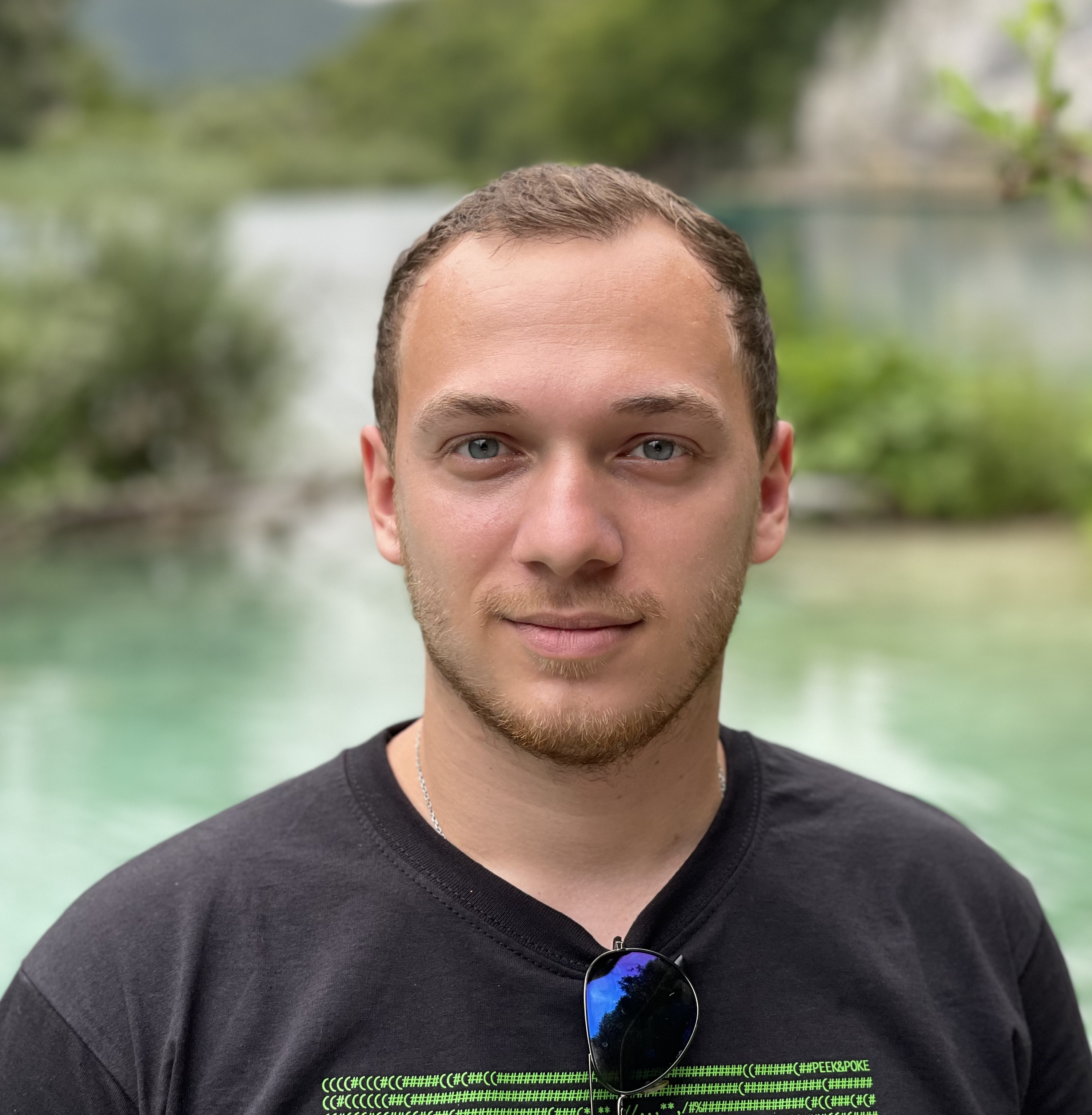}}]
{Nicolae-C\u{a}t\u{a}lin Ristea} graduated as valedictorian from the Faculty of Electronics, Telecommunications and Information Technology, University Politehnica of Bucharest, in 2019. He received the M.Sc.~degree in the image processing field and started the Ph.D.~at the same university. Nicolae is first author of multiple papers accepted at top-tier conferences and journals, such as CVPR and INTERSPEECH. His research interests include AI, computer vision, machine learning, signal processing and deep learning.
\end{IEEEbiography}

%\vspace{-1cm}
\begin{IEEEbiography}[{\includegraphics[width=0.6in,height=0.78in,clip,keepaspectratio]{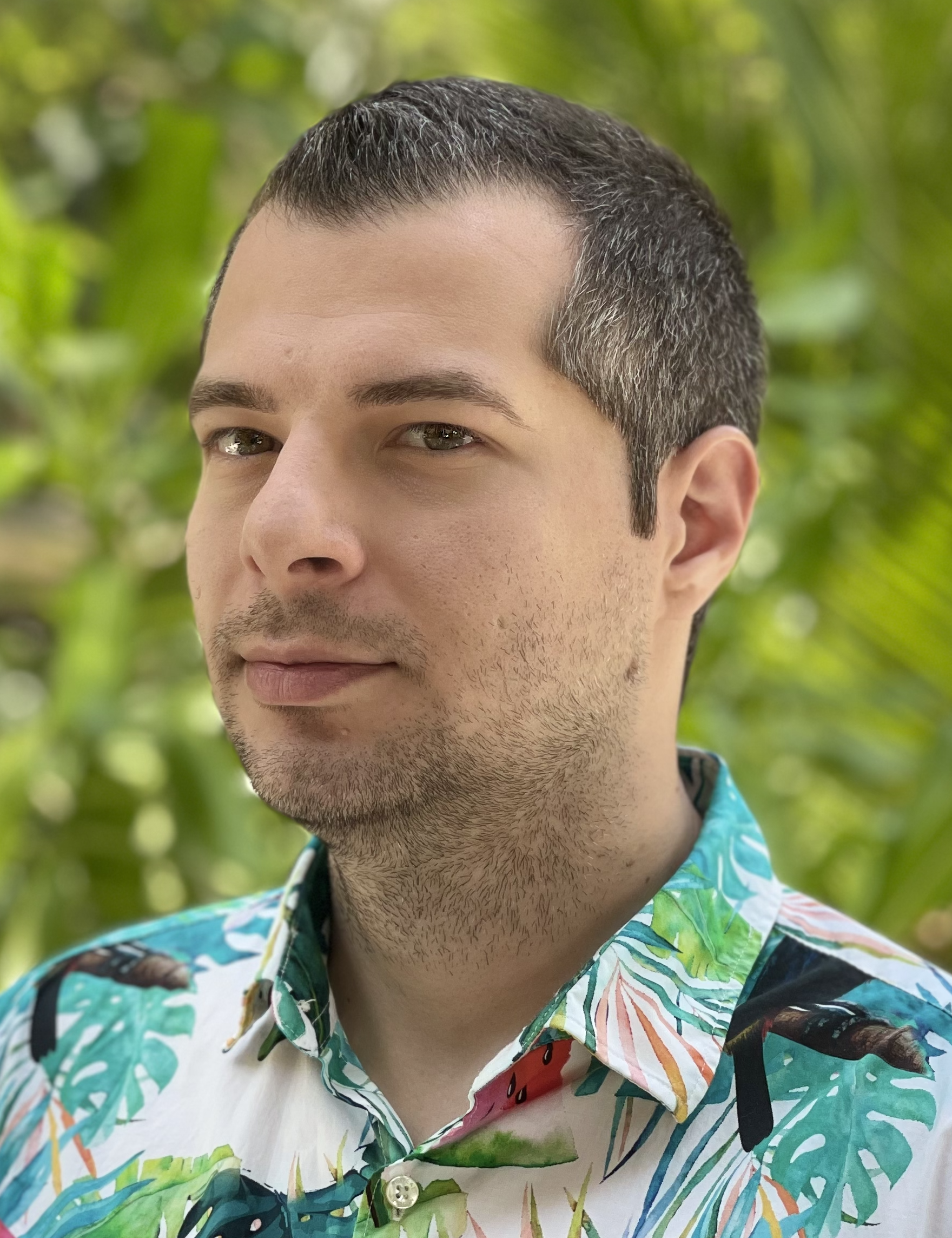}}]
{Radu Ionescu} is professor at the University of Bucharest, Romania. He completed his Ph.D.~at the University of Bucharest in 2013, receiving the 2014 Award for Outstanding Doctoral Research from the Romanian Ad Astra Association.
His research interests include machine learning, computer vision, image processing, computational linguistics and medical imaging. He published over 100 articles at international venues (including CVPR, NeurIPS, ICCV, ACL, SIGIR, EMNLP, NAACL, TPAMI, IJCV, CVIU), and a research monograph with Springer. Radu received the ``Caianiello Best Young Paper Award'' at ICIAP 2013. % for the paper entitled ``Kernels for Visual Words Histograms''. 
% Radu also received the 2017 ``Young Researchers in Science and Engineering'' Prize for young Romanian researchers and the ``Danubius Young Scientist Award 2018 for Romania''. % He participated at several international competitions obtaining top ranks: 4th place in the Facial Expression Recognition Challenge of WREPL 2013, 3rd place in the Native Language Identification Shared Task of BEA-8 2013, 2nd place in the Arabic Dialect Identification Shared Task of VarDial 2016, 
% 1st place in the Arabic Dialect Identification Shared Tasks of VarDial 2017 and 2018, 1st place in the Native Language Identification Shared Task of BEA-12 2017.
\end{IEEEbiography}

\begin{IEEEbiography}[{\includegraphics[width=0.6in,height=0.78in,clip,keepaspectratio]{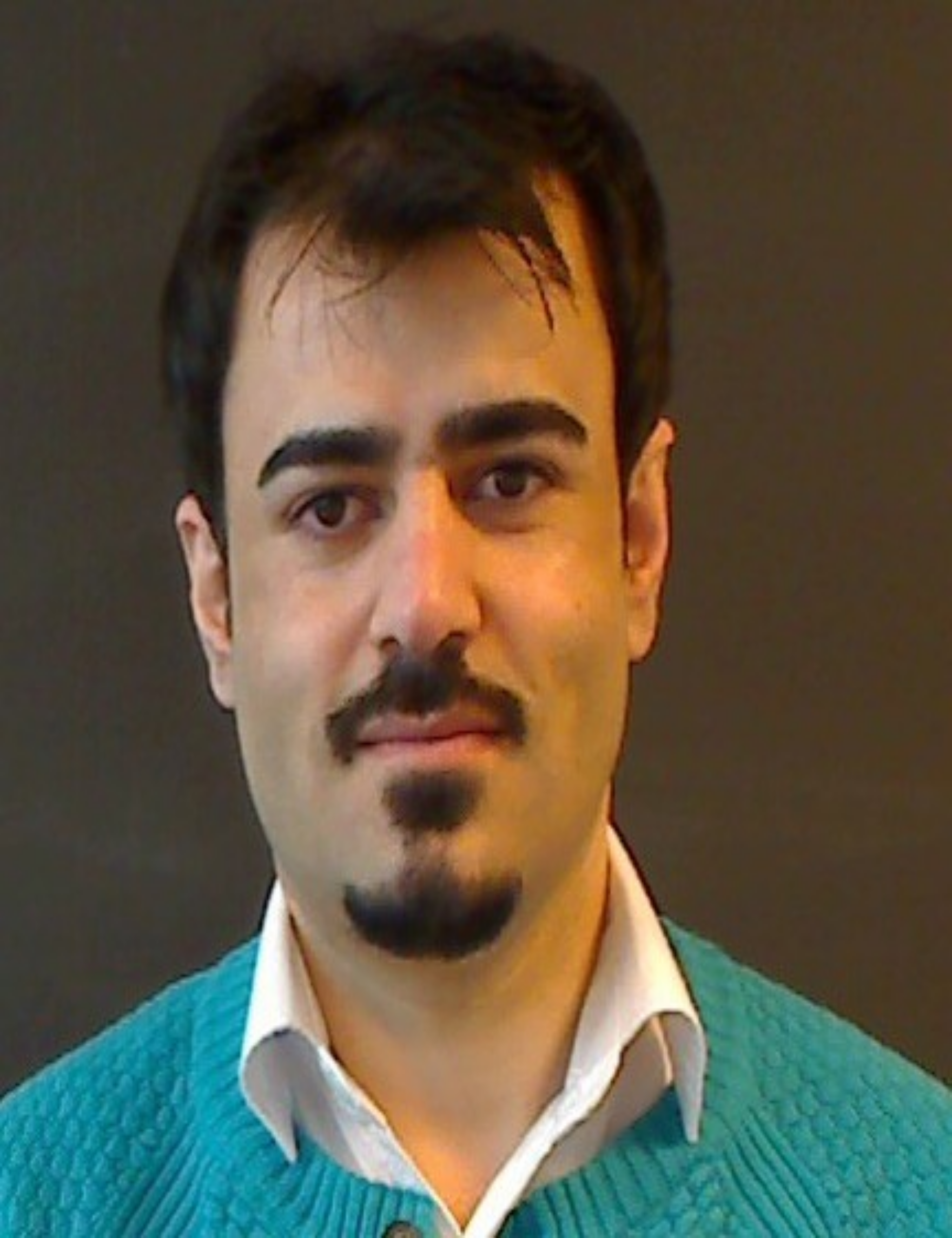}}]
{Kamal Nasrollahi} is working in a dual position, Professor of Computer Vision and Machine Learning at Aalborg University and Head of Machine Learning at Milestone Systems. He is interested in fair, ethical, and responsible use of technology, specifically machine learning applied to computer vision for topics like object detection, tracking, anomaly detection, and super-resolution.
\end{IEEEbiography}

%\vspace{-1cm}
\begin{IEEEbiography}[{\includegraphics[width=0.6in,height=0.78in,clip,keepaspectratio]{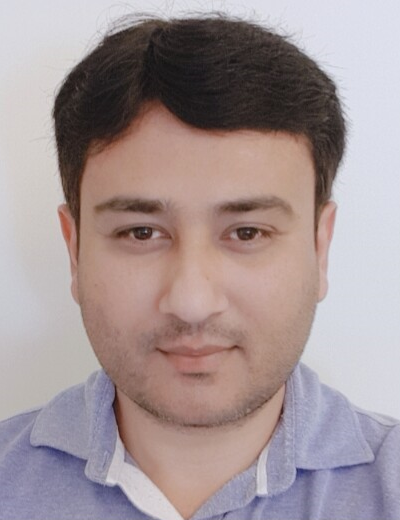}}]{Fahad Khan} is a faculty member at MBZ University of AI (MBZUAI), UAE and Link\"{o}ping University, Sweden. Prior to joining MBZUAI, he worked as a Lead Scientist at the Inception Institute of Artificial Intelligence (IIAI), UAE. He received the M.Sc.~degree in Intelligent Systems Design from Chalmers University of Technology, Sweden and a Ph.D.~degree in Computer Vision from Autonomous University of Barcelona, Spain. %He has achieved top ranks on various international challenges (Visual Object Tracking VOT: 1st 2014, 2016 and 2018, 2nd 2015; VOT-TIR: 1st 2015 and 2016; OpenCV Tracking: 1st 2015; PASCAL VOC: 1st 2010) and a best paper award at ICPR 2016. 
His research interests include a wide range of topics within computer vision, such as object recognition, object detection, action recognition and visual tracking. He has published articles in high-impact computer vision journals and conferences in these areas.
\end{IEEEbiography}

%\vspace{-1cm}
\begin{IEEEbiography}[{\includegraphics[width=0.6in,height=0.78in,clip,keepaspectratio]{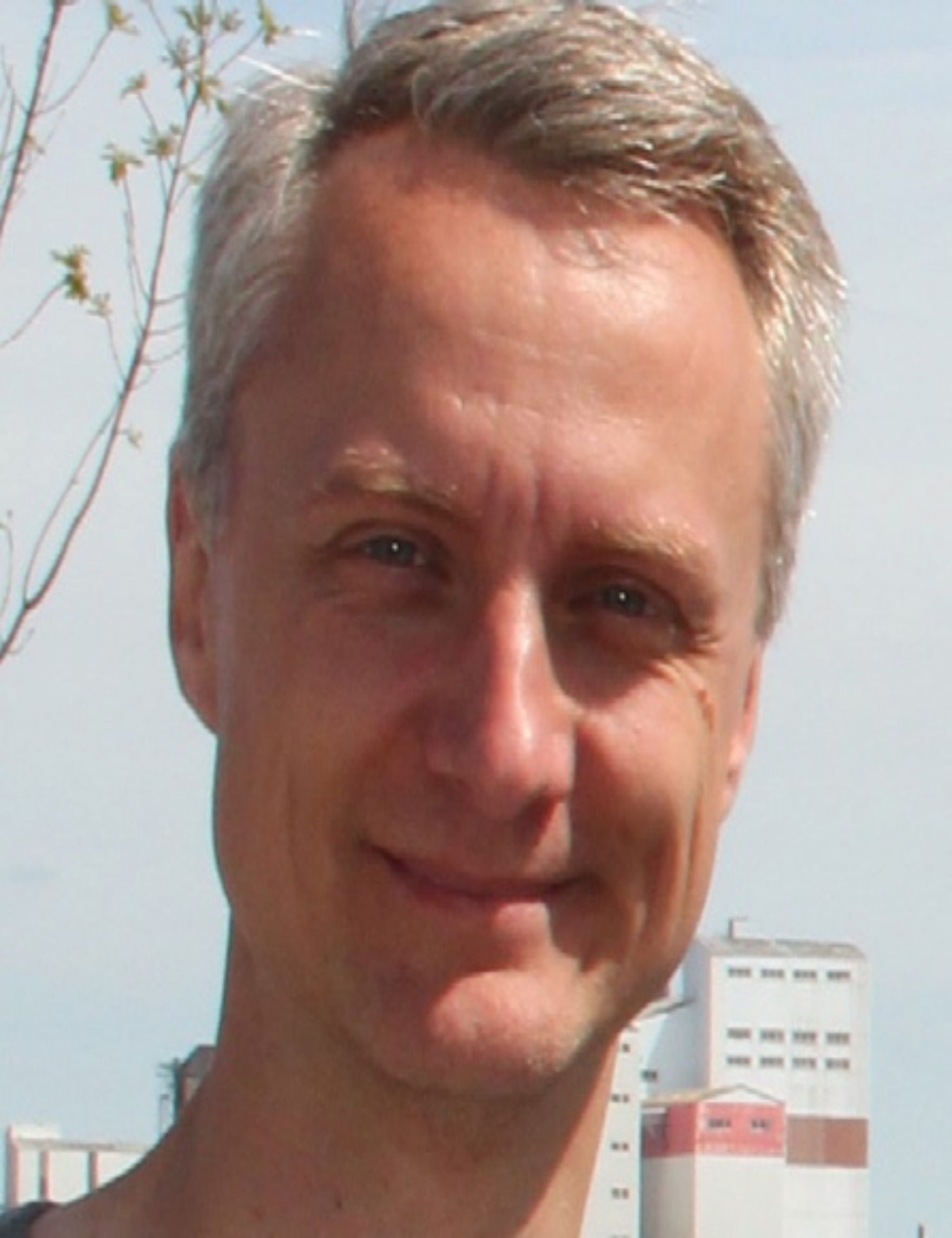}}]
{Thomas B. Moeslund} leads the Visual Analysis and Perception lab at Aalborg University, the Media Technology section at Aalborg University and the AI for the People Center at Aalborg University. His research covers all aspects of software systems for automatic analysis of visual data, especially including people.

\end{IEEEbiography}

% biography section
% 
% If you have an EPS/PDF photo (graphicx package needed) extra braces are
% needed around the contents of the optional argument to biography to prevent
% the LaTeX parser from getting confused when it sees the complicated
% \includegraphics command within an optional argument. (You could create
% your own custom macro containing the \includegraphics command to make things
% simpler here.)
%\vspace{-1cm}
\begin{IEEEbiography}[{\includegraphics[width=0.6in,height=0.78in,clip,keepaspectratio]{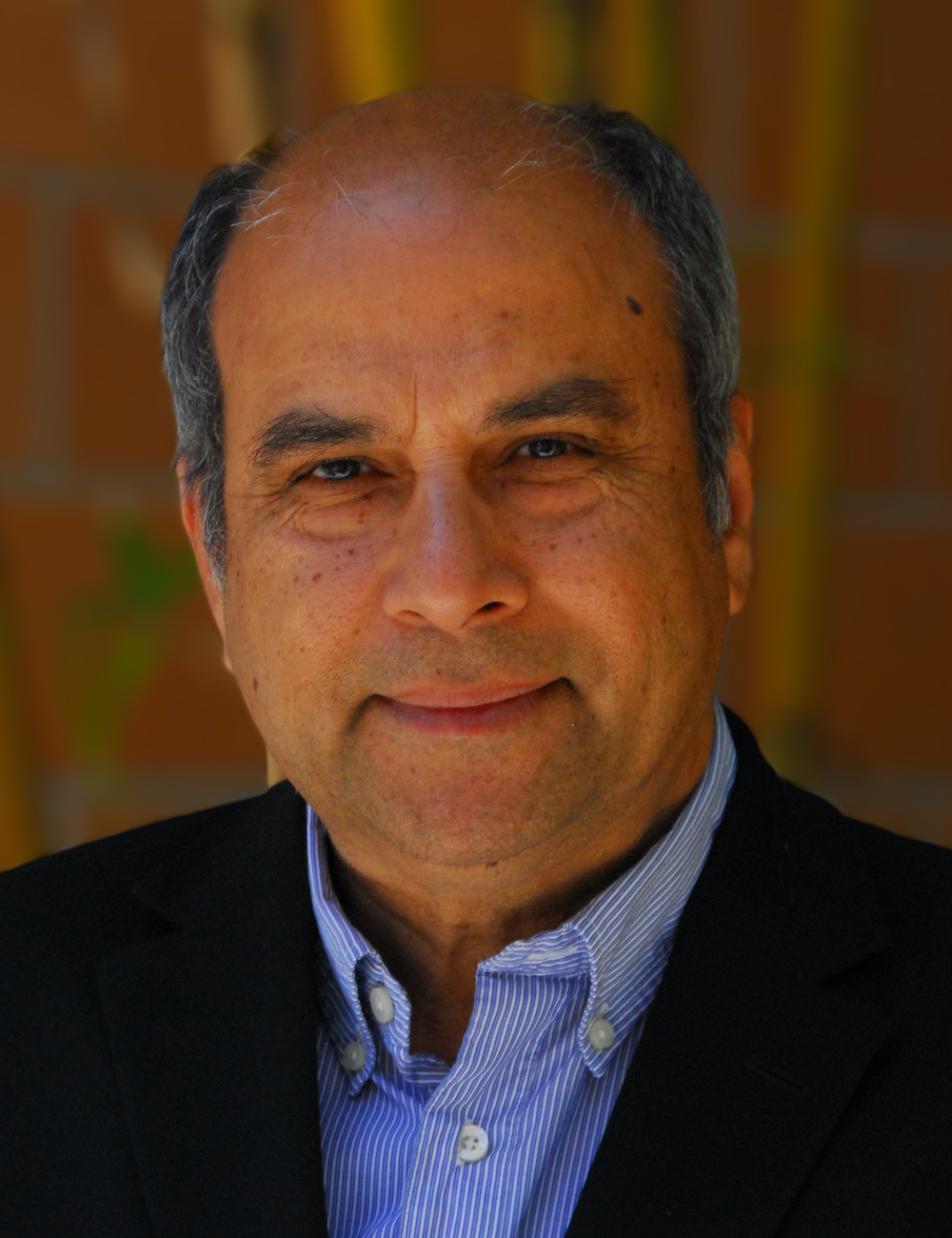}}]
{Mubarak Shah}, the UCF Trustee chair professor, is the founding director of the Center for Research in Computer Vision at the University of Central Florida (UCF). He is a fellow of the NAI, IEEE, AAAS, IAPR and SPIE. % He is an editor of an international book series on video computing, was editor-in-chief of Machine Vision and Applications and an associate editor of ACM Computing Surveys and IEEE T-PAMI. He was the program co-chair of CVPR 2008, an associate editor of the IEEE T-PAMI and a guest editor of the special issue of the International Journal of Computer Vision on Video Computing. 
His research interests include video surveillance, visual tracking, human activity recognition, visual analysis of crowded scenes, video registration, UAV video analysis, among others. He has served as an ACM distinguished speaker and IEEE distinguished visitor speaker. He is a recipient of ACM SIGMM Technical Achievement award; IEEE Outstanding Engineering Educator Award; Harris Corporation Engineering Achievement Award; an honorable mention for the ICCV 2005 ``Where Am I?'' Challenge Problem; 2013 NGA Best Research Poster Presentation; 2nd place in Grand Challenge at the ACM Multimedia 2013 conference; and runner up for the best paper award in ACM Multimedia Conference in 2005 and 2010. At UCF he has received the Pegasus Professor Award; University Distinguished Research Award; Faculty Excellence in Mentoring Doctoral Students; Scholarship of Teaching and Learning award; Teaching Incentive Program award; Research Incentive Award.
\end{IEEEbiography}

% that's all folks
\end{document}